\pdfoutput=1
\def\arxivversion{}

\documentclass[10pt,twocolumn,letterpaper]{article}

\ifdefined\arxivversion
\usepackage[pagenumbers]{cvpr}
\makeatletter
\let\cvpr@maketitle\@maketitle
\def\@maketitle{%
  \setbox\z@\vbox{\def\@author{Anonymous \confName~submission}\cvpr@maketitle}%
  \vbox to\ht\z@{%
    \null\vskip 10pt
    \begin{center}
      {\Large\bf\@title\par}
      \vspace*{12pt}
      {\large\lineskip.5em\begin{tabular}[t]{c}\@author\end{tabular}\par}
    \end{center}
    \vss}}
\makeatother
\else
\usepackage[review]{cvpr}
\fi

\definecolor{cvprblue}{rgb}{0.21,0.49,0.74}

\def\confName{CVPR}
\def\confYear{2027}
\usepackage{url}
\usepackage{booktabs}
\usepackage{amsfonts}
\usepackage{nicefrac}
\usepackage{microtype}
\usepackage{xcolor}
\usepackage{lipsum}

\definecolor{dblue}{RGB}{52, 104, 192}
\definecolor{dgreen}{RGB}{65, 171, 93}
\definecolor{dred}{RGB}{210, 69, 69}
\definecolor{dpurple}{RGB}{148, 103, 189}
\definecolor{dorange}{RGB}{255, 127, 14}
\definecolor{dcyan}{RGB}{35, 70, 100}
\definecolor{dviolet}{rgb}{0.40, 0.22, 0.60}
\definecolor{meshgrey}{RGB}{140, 143, 147}
\definecolor{meshorange}{RGB}{222, 140, 52}
\definecolor{contactred}{RGB}{213, 71, 81}

\definecolor{color_cem}{RGB}{102, 194, 165}
\definecolor{color_ecbm}{RGB}{252, 141, 98}
\definecolor{color_ar}{RGB}{141, 160, 203}
\definecolor{color_scbm}{RGB}{231, 138, 195}

\usepackage[most]{tcolorbox}
\newtcolorbox{questionbox}{
  enhanced, frame hidden, sharp corners, boxrule=0pt,
  colback=black!4,
  borderline west={2pt}{0pt}{cvprblue},
  left=8pt, right=8pt, top=5pt, bottom=5pt, boxsep=0pt,
  before skip=7pt plus 1pt, after skip=7pt plus 1pt,
  fontupper=\itshape, halign=flush center
}

\usepackage{bbm}
\usepackage{graphicx}
\usepackage{amsmath}
\usepackage{amssymb}
\usepackage{amsthm}
\usepackage{multirow}
\usepackage{caption}
\usepackage{wrapfig}
\usepackage{placeins}
\usepackage{enumitem}
\usepackage[colorlinks,
            pagebackref,
            breaklinks,
            linkcolor=red,
            anchorcolor=magenta,
            urlcolor=dblue,
            citecolor=dblue
            ]{hyperref}
\usepackage{subcaption}
\usepackage{multicol}
\usepackage{colortbl}
\usepackage{xspace}
\usepackage{siunitx}
\usepackage{array}
\usepackage{pifont}

\definecolor{tabgray}{HTML}{f2f2f2}
\definecolor{tablavender}{HTML}{E6F7E6}
\definecolor{tabnavy}{HTML}{00008B}
\definecolor{tabwhite}{HTML}{FFFFFF}
\definecolor{linf}{HTML}{8172b2}
\definecolor{l2}{HTML}{ccb974}
\definecolor{l1}{HTML}{64b5cd}

\definecolor{lavender}{HTML}{9ad756}
\definecolor{DBlue}{RGB}{31, 119, 180}
\definecolor{DGreen}{RGB}{44,160,44}
\definecolor{DRed}{RGB}{214,39,40}
\definecolor{dorange}{RGB}{255, 127, 14}
\definecolor{dorange2}{RGB}{255, 165, 0}

\newcommand{\method}{\textsc{Horcrux}\xspace}
\newcommand{\methodall}{\textsc{Horcrux}\textsuperscript{$\ddagger$}\xspace}

\newcommand{\na}{\textcolor{black!55}{N/A}}

\newcommand{\cmb}{\textsuperscript{$\dagger$}}

\usepackage{bm, dsfont}

\let\hat\widehat
\let\tilde\widetilde

\newcommand{\RR}{\mathbb{R}}

\newcommand{\argmin}{\mathop{\mathrm{argmin}}}

\makeatletter
\newcommand{\ve}{\@ifnextchar\bgroup{\velong}{{\bm{e}}}}
\newcommand{\velong}[1]{{\bm{#1}}}
\makeatother

\title{Generalizable Feed-Forward 3D Hand-Object Reconstruction\\ with Coarse-to-Fine Contact Estimation}

\ifdefined\arxivversion
\author{Seonghwan Park\qquad Yong-Hoon Kwon\qquad Min-Gyu Park\\
Korea Electronics Technology Institute (KETI)\\
{\tt\small \{shpark97, yhkwon, mpark\}@keti.re.kr}
}
\fi

\begin{document}
\twocolumn[{%
\renewcommand\twocolumn[1][]{#1}%
\maketitle\vspace{-2.05em}%
\begin{center}
\includegraphics[width=\linewidth]{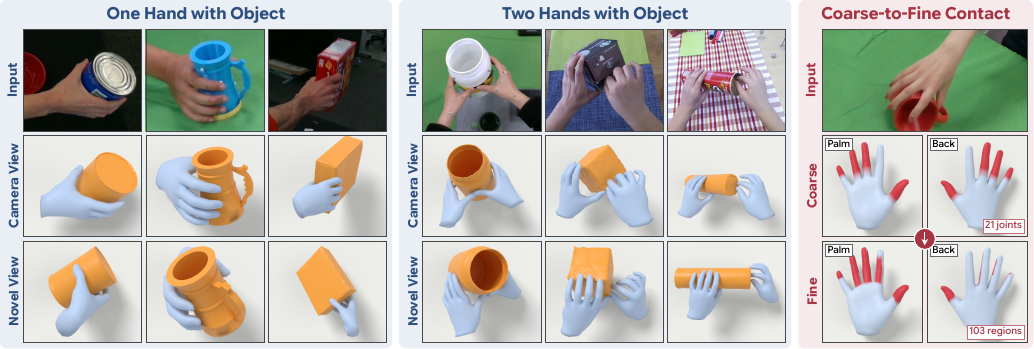}
\par\vspace{-0.65em}
\captionof{figure}{We present \method, a unified feed-forward transformer that jointly reconstructs hands and the manipulated object from a single image and estimates their contact. A single model handles one hand (left) and two hands (middle) across datasets. Coarse per-joint contact (right) is resolved over finer hand regions, which localizes it on the palm rather than the back.}
\label{fig:teaser}
\end{center}%
\vspace{0.4em}
}]
\begin{abstract}
Reconstructing the 3D poses of hands and objects from a single image, together with the contact between them, is fundamental to understanding manipulation.
To be practical, the reconstruction must generalize to unseen objects, and the contact must be localized precisely.
Yet joint feed-forward estimators often rely on predefined object sets, while contact is typically predicted from image features alone, without the explicit geometry needed to localize it.
To address both limitations, we present \method, a unified feed-forward transformer with coarse-to-fine contact estimation.
For unseen objects, a frozen foundation model encodes the object from a given or image-reconstructed mesh rather than from object-specific knowledge.
A shared transformer then reasons jointly over the object and one or two hands to keep them aligned.
For precise contact, coarse per-joint predictions from latent features are resolved over finer hand regions using the reconstructed geometry.
Across seven benchmarks, \method achieves state-of-the-art object and hand pose accuracy among joint estimators and outperforms a dedicated contact estimator in fine-grained contact.
These results suggest that contact is best estimated jointly with reconstruction, whose geometry directly localizes it.
\looseness=-1\par
\end{abstract}
    
\addtocontents{toc}{\protect\setcounter{tocdepth}{0}}
\section{Introduction}
\label{sec:introduction}

Understanding how hands interact with objects is central to robot learning from human video~\citep{qin2022dexmv, sivakumar2022telekinesis}, augmented reality~\citep{banerjee2025hot3d}, and interaction analysis~\citep{shan2020understanding}.
This requires recovering the 3D poses and relative arrangement of the hands and object from a monocular image~\citep{hasson2019obman, tse2022collaborative, qi2024hoisdf}.
Beyond pose, contact reveals where the hands touch the object and thus how it is held~\citep{yang2021cpf}.
In practice, these applications involve everyday objects and depend on exactly how each is grasped.
The reconstruction must therefore generalize to objects unseen in training, and the contact must be localized precisely on the hand surface.
Yet both remain challenging, as models have no prior knowledge of an unseen object, and contact predicted from image features alone is often imprecise.

Existing methods address these requirements only in part.
Joint feed-forward estimators~\citep{liu2021semihand, lin2023hfl, wang2025unihope} reason about the hand and object together but rely on knowledge of predefined objects, so they generalize poorly to unseen ones.
Composing per-entity foundation models~\citep{pavlakos2024hamer, chen2026posegam, liu2025easyhoi} handles unseen objects given their meshes, yet estimating each pose independently can misalign the hand and object.
Meanwhile, reconstruction methods mainly use contact as an auxiliary signal (\eg, for supervision~\citep{bansal2026hopformer} or to refine hand pose~\citep{yang2021cpf, tse2022s2contact} and object shape~\citep{hu2024choi}), while dedicated estimators~\citep{narasimhaswamy2020contacthands, jung2025haco, jung2026contactprompt} predict it without reconstruction.
Even those predicting dense contact~\citep{jung2025haco, jung2026contactprompt} rely on the image alone, without the explicit hand-object geometry needed to localize it precisely or to keep it consistent with the reconstruction.
As a result, no existing feed-forward model handles unseen objects while also localizing contact precisely on the hand surface.

Achieving both means combining what existing methods offer only separately.
To generalize to unseen objects without misalignment, a model should encode the object from its geometry rather than from object-specific knowledge, while still reasoning about the hand and object together.
Precise contact similarly calls for two complementary cues.
Latent features indicate which joints are likely in contact but lack the explicit surface geometry needed to resolve which side of a finger touches the object.
Contact inferred from reconstructed meshes captures this geometry but inherits their pose errors.
A model that reconstructs the hand and object has access to both cues, and combining them can yield finer and more accurate contact.

Building on these insights, we propose \method, a unified feed-forward transformer that performs \underline{H}and-\underline{O}bject \underline{R}econstruction with \underline{C}ontact estimation and \underline{R}obust generalization to \underline{U}nseen objects.
It rests on two components: (i)~\textbf{joint reasoning over foundation features} for unseen objects and (ii)~\textbf{coarse-to-fine contact estimation} for precise contact.
The former encodes the object with a frozen object-pose foundation model~\citep{chen2026posegam}, whose features capture its geometry from a given or image-reconstructed mesh~\citep{sam3d2025, hunyuan3d2025v21}.
A shared transformer then reasons jointly over these features and those of each hand to predict the hand and object poses, whether one or two hands are present.
The latter in turn draws on the resulting reconstruction rather than the image alone.
A coarse head first predicts per-joint contact from the latent features of this transformer, which keeps it informative even under imprecise reconstruction.
A fine head then resolves it over finer hand regions~\citep{jung2026contactprompt} using distances to the reconstructed object, separating the palm side of each finger from its back.
Contact is thereby localized precisely without a separate estimator.
Figure~\ref{fig:teaser} summarizes the outputs of~\method.

Our key contributions are summarized as follows:
\begin{itemize}[leftmargin=1em, topsep=-\parskip, itemsep=0.1em]
    \item We propose \method, a unified feed-forward transformer that reasons jointly over frozen foundation features, generalizing to unseen objects with given or image-reconstructed meshes and handling one or two hands.

    \item We introduce coarse-to-fine contact estimation, which predicts per-joint contact from latent features and resolves it over finer hand regions using the reconstructed geometry, without a separate estimator.

    \item Across seven benchmarks, \method achieves state-of-the-art object and hand pose accuracy among joint estimators. It also outperforms a dedicated estimator in fine-grained contact.
\end{itemize}

\section{Related Work}
\label{sec:related_work}

\begin{figure*}[t]
\centering
\includegraphics[width=\linewidth]{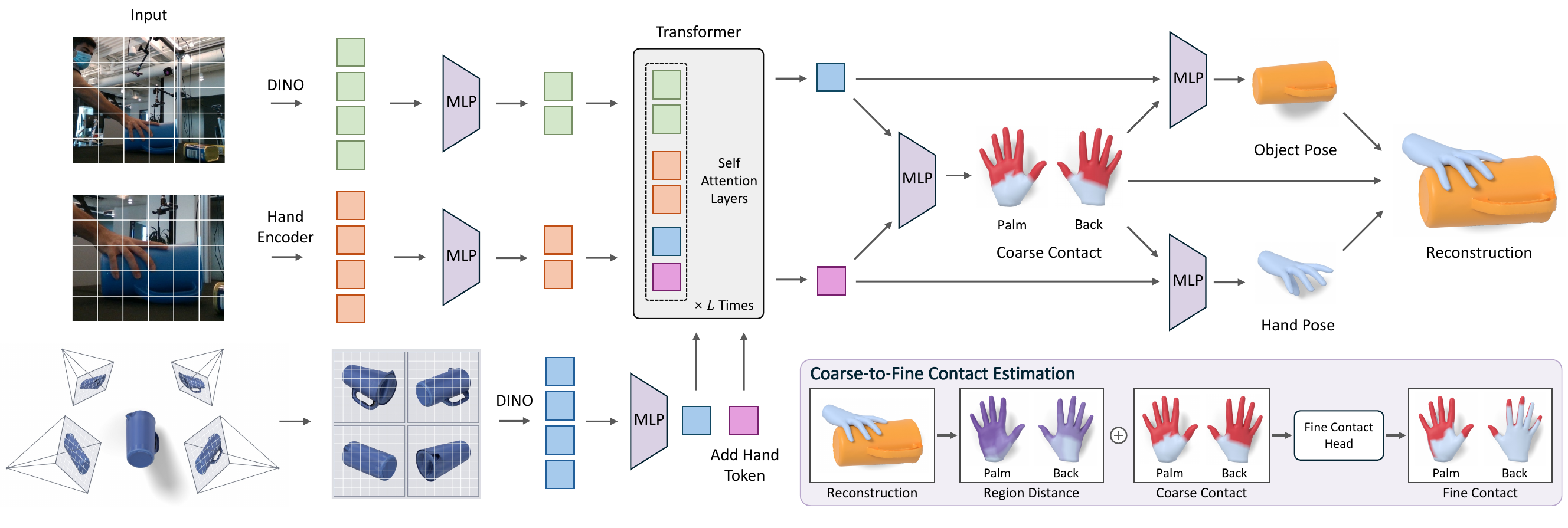}
\par\vspace{-0.5em}
\caption{Overview of \method, which consists of (i)~\textit{joint reasoning over foundation features} and (ii)~\textit{coarse-to-fine contact estimation}. With frozen encoders, the former tokenizes the image and each hand crop and encodes the object from its mesh, so the model is not tied to its training objects. A shared transformer reasons jointly over these tokens to keep the hands and object aligned, and heads decode their poses and coarse per-joint contact. The latter resolves this contact over finer hand regions using region distances from the reconstruction.}
\label{fig:method}
\end{figure*}

\paragraph{Joint hand-object reconstruction.}
Feed-forward methods reconstruct the hand and object jointly from a single image~\citep{tekin2019ho, doosti2020hope, hasson2020leveraging, hampali2022keypoint, liu2021semihand, aboukhadra2023thor, bansal2026hopformer, yang2026homae, zhou2025vpho}, often coupling them for consistency through shared features or signed distance fields~\citep{tse2022collaborative, lin2023hfl, ihoi, chen2022alignsdf, chen2023gsdf, qi2024hoisdf}.
Yet their generalization is limited, as many handle only predefined object sets and rely on the appearance of their training objects~\citep{liu2021semihand, lin2023hfl, wang2025unihope}.
For unseen objects, some learn to reconstruct the shape from the image~\citep{ihoi, chen2022alignsdf, chen2023gsdf} or use semantic and point-cloud cues~\citep{yang2026genhoi}.
\method instead takes the shape as input (\ie, a given or image-reconstructed mesh~\citep{sam3d2025, hunyuan3d2025v21}) and encodes it with a frozen object-pose foundation model~\citep{chen2026posegam}, relying on geometry rather than on the appearance of training objects.

\paragraph{Per-entity foundation models.}
Foundation models generalize broadly, but each addresses the hand or the object alone.
Hand regressors recover MANO~\citep{romero2017mano} hands from in-the-wild images~\citep{pavlakos2024hamer, potamias2024wilor} or egocentric videos~\citep{zhang2025hawor}.
Zero-shot object-pose estimators handle unseen objects given their meshes~\citep{labbe2022megapose, ausserlechner2024zs6d, ornek2024foundpose, nguyen2024gigapose, chen2026posegam}, typically by comparing rendered views with the image, so their internal features encode the object geometry.
To handle both, recent systems build on pretrained visual backbones~\citep{oquab2024dinov2, simeoni2025dinov3, darcet2024registers, wang2025vggt} for hand-held object reconstruction~\citep{zhang2024moho} or combine per-entity estimates through per-image test-time optimization~\citep{liu2025easyhoi}.
However, independently estimated poses can misalign the hand and object, and the object pose degrades under hand occlusion.
\method uses the features of foundation models rather than their outputs, training a shared transformer to reason jointly over them and predict the hand and object poses.

\paragraph{Hand-object contact estimation.}
Reconstruction methods mainly use contact as an auxiliary signal for supervision~\citep{hasson2019obman, bansal2026hopformer}, test-time refinement~\citep{grady2021contactopt, tse2022s2contact}, and evaluation~\citep{fan2023arctic}. Some further predict it to refine hand pose~\citep{yang2021cpf}, human-object reconstruction~\citep{nam2024contho, xie2022chore}, and object shape~\citep{hu2024choi}.
As an output, contact comes from one of two cues.
Geometry-based methods derive it from the reconstruction, as in hand-object distance fields~\citep{karunratanakul2020grasping, chen2023gsdf, qi2024hoisdf} or contact maps for interaction recognition~\citep{cho2023h2otr}, but inherit its errors.
Image-based methods instead predict it directly, ranging from coarse per-hand contact states~\citep{shan2020understanding, narasimhaswamy2020contacthands} to dense contact on human bodies~\citep{tripathi2023deco} and hand surfaces~\citep{jung2025haco}.
Recent work~\citep{jung2026contactprompt} reasons over finer hand regions by prompting multimodal models (\eg, GPT~\citep{openai2026gpt55} and Claude~\citep{anthropic2026claudeopus47}).
However, these methods rely on image features alone and are decoupled from reconstruction, so their contact is often imprecise without explicit geometry and may contradict the recovered poses.
\mbox{\method} instead combines both cues within reconstruction, predicting coarse per-joint contact from latent features and resolving it over finer hand regions using the reconstructed geometry.

\section{Method}
\label{sec:method}

We present \method, a unified feed-forward transformer that reconstructs the hands and the manipulated object from a monocular image and estimates their fine-grained contact.
After defining the problem (Section~\ref{subsec:setup}), we describe how a shared transformer reasons jointly over frozen foundation features to predict the poses and coarse contact (Section~\ref{subsec:arch}).
We then detail how coarse-to-fine contact estimation resolves it using the resulting reconstruction (Section~\ref{subsec:heads}) and how the model is trained (Section~\ref{subsec:loss}).
Figure~\ref{fig:method} shows the pipeline.

\subsection{Problem Definition}
\label{subsec:setup}
Given a monocular RGB image $I$, a crop $I_h$ of each detected hand $h$ (left or right)~\citep{pavlakos2024hamer}, and a 3D mesh $V_o\in\RR^{N_o\times 3}$ of the manipulated object, we learn the mapping
\begin{equation}
f\big(I,\{I_h\},V_o\big)=\big(\{\theta_h,\beta_h,\mathbf{t}_h,\mathbf{c}_h\},\ \mathbf{R}_o,\mathbf{t}_o\big),
\label{eq:task}
\end{equation}
where $\theta_h\in\RR^{48}$ and $\beta_h\in\RR^{10}$ are the MANO pose and shape~\citep{romero2017mano} of hand $h$, $\mathbf{t}_h\in\RR^{3}$ is its translation, $\mathbf{R}_o\in\mathbb{SO}(3)$ and $\mathbf{t}_o\in\RR^{3}$ are the rotation and translation of the object, and $\mathbf{c}_h$ is the contact between hand $h$ and the object.
All poses are expressed in the camera frame.
Here, $\{\cdot\}$ denotes the set over the one or two hands in the image.

\subsection{Joint Reasoning over Foundation Features}
\label{subsec:arch}
We realize $f$ by tokenizing the inputs with frozen foundation encoders and reasoning jointly over the tokens.
A DINOv3 encoder $E_{\mathrm{img}}$~\citep{simeoni2025dinov3} extracts scene tokens from $I$, and a hand encoder $E_h$~\citep{pavlakos2024hamer} maps each $I_h$ to crop tokens:
\begin{equation}
\mathbf{Z}_{\mathrm{img}}=\psi_{\mathrm{img}}\big(E_{\mathrm{img}}(I)\big),\qquad \mathbf{Z}_h = \psi_h\big(E_h(I_h)\big),
\label{eq:fuse}
\end{equation}
where $\psi_{\mathrm{img}}$ and $\psi_h$ are learnable projections, the latter shared across hands.

For the object, we encode its 3D geometry rather than learning an instance embedding.
A frozen object-pose foundation model $E_o$~\citep{chen2026posegam} renders reference views of $V_o$ and aggregates them with $I$.
Its decoded pose degrades when the grasping hand occludes the object (see Supp.~\ref{app:analyses}), but its internal features still encode the object geometry.
We map these features into an object token with a learnable projection $\psi_o$:
\begin{equation}
\mathbf{z}_o = \psi_o\big(E_o(I,V_o)\big).
\label{eq:inject}
\end{equation}
The model is thus not tied to its training objects and can use approximate image-reconstructed meshes (see Section~\ref{subsec:domaingen}).

A shared transformer $\mathcal{F}$ then reasons jointly over these tokens to keep the hands and the object aligned.
With a learnable token $\mathbf{z}_h$ per hand, the tokens form the sequence
\begin{equation}
\mathbf{Z}=\big[\,\mathbf{Z}_{\mathrm{img}};\ \{\mathbf{Z}_h\};\ \mathbf{z}_o;\ \{\mathbf{z}_h\}\,\big].
\label{eq:agg}
\end{equation}
The transformer updates this sequence with $L$ global self-attention blocks~\citep{wang2025vggt}, in which every token attends to all others.
A missing hand contributes neither tokens nor outputs, so the same transformer serves one- and two-hand configurations.
We denote the updated tokens of hand $h$ by $\hat{\mathbf{z}}_h$ and $\hat{\mathbf{Z}}_h$, and that of the object by $\hat{\mathbf{z}}_o$.

\begin{figure}[t]
\centering
\begin{subfigure}{\linewidth}
\centering
\includegraphics[width=\linewidth]{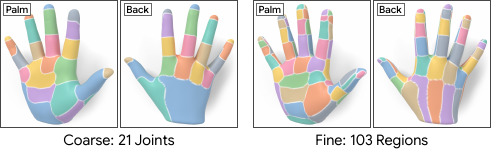}
\caption{Coarse and fine contact units}
\label{fig:parts_units}
\end{subfigure}\\[3pt]
\begin{subfigure}{\linewidth}
\centering
\includegraphics[width=\linewidth]{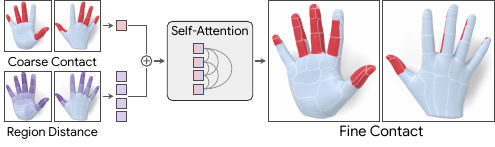}
\caption{Fine contact head}
\label{fig:parts_example}
\end{subfigure}
\par\vspace{-0.5em}
\caption{Coarse-to-fine contact estimation. (a) Per-joint contact cannot separate the palmar and dorsal sides of a finger, whereas finer hand regions~\citep{jung2026contactprompt} can. (b) In the fine head, the coarse contact identifies the touching joints, and region distances from the reconstruction (\textcolor{dviolet}{violet}, darker when closer) localize it on the palmar side.}
\label{fig:parts}
\end{figure}

Lightweight heads~\citep{pavlakos2024hamer} then decode these tokens into the outputs of Eqn.~\eqref{eq:task}.
A hand head shared across hands predicts $\theta_h$, $\beta_h$, and $\mathbf{t}_h$ from $\hat{\mathbf{z}}_h$.
From $\hat{\mathbf{z}}_o$, an object head predicts the rotation $\mathbf{R}_o$ and the translation offset $\Delta\mathbf{t}_o$ from a grasping hand $a$, giving $\mathbf{t}_o=\mathbf{t}_a+\Delta\mathbf{t}_o$.
This keeps the hand and object aligned even when the absolute hand translation is inaccurate.
A coarse contact head further predicts per-joint contact $\hat{\mathbf{c}}_h\in[0,1]^{J}$ over $J=21$ joints from $\hat{\mathbf{Z}}_h$ and $\hat{\mathbf{z}}_o$:
\begin{equation}
\hat{\mathbf{c}}_h=\sigma\big(\phi_{\mathrm{c}}(\hat{\mathbf{Z}}_h^{\mathrm{pool}}\oplus\hat{\mathbf{z}}_o)\big),
\label{eq:contact}
\end{equation}
where $\hat{\mathbf{Z}}_h^{\mathrm{pool}}$ is $\hat{\mathbf{Z}}_h$ attention-pooled with a learned query, $\oplus$ is concatenation, and $\phi_{\mathrm{c}}$ is an MLP shared across hands.

\subsection{Coarse-to-Fine Contact Estimation}
\label{subsec:heads}
\label{subsec:cbl}
The coarse contact $\hat{\mathbf{c}}_h$ of Eqn.~\eqref{eq:contact} is predicted from latent tokens rather than measured on the reconstructed meshes, so it does not inherit their pose errors.
At the joint level, however, it cannot determine which side of a finger touches the object (Figure~\ref{fig:parts}a), and neither can coarser per-hand contact states~\citep{shan2020understanding, narasimhaswamy2020contacthands}.
We therefore resolve $\hat{\mathbf{c}}_h$ over the $P=103$ finer hand regions~\citep{jung2026contactprompt} (\eg, the palmar and dorsal sides of each finger segment), giving $\tilde{\mathbf{c}}_h\in[0,1]^{P}$.
Whereas prior work prompts multimodal models (\eg, GPT~\citep{openai2026gpt55} and Claude~\citep{anthropic2026claudeopus47}) to predict region contact from the image alone, we localize it with the reconstructed geometry that latent tokens lack.
Yet geometry alone inherits the pose errors that latent tokens avoid.
We thus combine the coarse contact with region distances from the reconstruction (Figure~\ref{fig:parts}b), so that the reconstruction itself becomes evidence for contact.

Each reconstructed hand vertex $\mathbf{v}_{h,i}$ is mapped into the object frame, where the signed distance function $\mathrm{SDF}_o$ of the object mesh~\citep{park2019deepsdf} gives its distance
\begin{equation}
d_{h,i}=\mathrm{SDF}_o\big(\mathbf{R}_o^{\top}(\mathbf{v}_{h,i}-\mathbf{t}_o)\big),
\label{eq:regiondist}
\end{equation}
and the region distance $\mathbf{u}_{h,p}$ summarizes the distances $d_{h,i}$ of the vertices in region $p$ (\eg, their minimum and mean).
The fine head $\mathcal{F}_{\mathrm{f}}$, a one-layer transformer~\citep{vaswani2017attention}, then takes the region tokens $\hat{\mathbf{c}}_h\oplus\mathbf{u}_{h,p}$ with learnable region embeddings and lets them interact through self-attention:
\begin{equation}
\tilde{\mathbf{c}}_h=\sigma\Big(\mathcal{F}_{\mathrm{f}}\big(\{\hat{\mathbf{c}}_h\oplus\mathbf{u}_{h,p}\}_{p=1}^{P}\big)\Big).
\label{eq:fine}
\end{equation}
The coarse contact thus identifies the touching joints, and the region distances localize it on the hand surface, so fine-grained contact needs no separate estimator.

\subsection{Training}
\label{subsec:loss}
\begin{table*}[t]
\centering
\caption{In-domain comparison on the DexYCB s0 and HO3Dv2 test splits. The best and second-best results are marked in \textbf{bold} and \underline{underlined}, respectively. Overall, \method\ leads every baseline in MJE and every joint estimator in object AUC. \cmb\ denotes a joint pose obtained by pairing the method with the MegaPose~\citep{labbe2022megapose} object (highest AUC) or the H2ONet~\citep{xu2023h2onet} hand (lowest PA-MJE).}
\label{tab:indomainpose}
\par\vspace{-0.5em}
\resizebox{\linewidth}{!}{%
\begin{tabular}{l|cccccccc|cccccccc}
\toprule
& \multicolumn{8}{c|}{\textbf{DexYCB}~\citep{chao2021dexycb}} & \multicolumn{8}{c}{\textbf{HO3D}~\citep{hampali2020honnotate}} \\
\cmidrule(lr){2-9}\cmidrule(lr){10-17}
& Object Pose & \multicolumn{4}{c}{Hand Pose} & \multicolumn{3}{c|}{Joint Pose} & Object Pose & \multicolumn{4}{c}{Hand Pose} & \multicolumn{3}{c}{Joint Pose} \\
\cmidrule(lr){2-2}\cmidrule(lr){3-6}\cmidrule(lr){7-9}\cmidrule(lr){10-10}\cmidrule(lr){11-14}\cmidrule(lr){15-17}
\textbf{Method} & AUC ($\uparrow$) & MJE ($\downarrow$) & PA-MJE ($\downarrow$) & VPE ($\downarrow$) & PA-VPE ($\downarrow$) & SD ($\downarrow$) & IV ($\downarrow$) & CDev ($\downarrow$) & AUC ($\uparrow$) & MJE ($\downarrow$) & PA-MJE ($\downarrow$) & VPE ($\downarrow$) & PA-VPE ($\downarrow$) & SD ($\downarrow$) & IV ($\downarrow$) & CDev ($\downarrow$) \\
\midrule
\textbf{\underline{Hand Pose Estimation}} & & & & & & & & & & & & & & & & \\
HandOccNet~\citep{park2022handoccnet} & \na & 12.79 & 6.15 & 12.39 & 5.95 & 4.53\cmb & 16.32\cmb & 42.45\cmb & \na & 28.94 & 10.26 & 28.10 & 10.21 & \underline{5.41}\cmb & 22.51\cmb & 16.49\cmb \\
MobRecon~\citep{chen2022mobrecon} & \na & 14.17 & 6.85 & 13.16 & 5.97 & 4.64\cmb & 16.83\cmb & 42.73\cmb & \na & 29.61 & 10.47 & 28.65 & 10.76 & 5.66\cmb & 19.07\cmb & 18.07\cmb \\
H2ONet~\citep{xu2023h2onet} & \na & 13.76 & 5.73 & 13.44 & 5.84 & 4.68\cmb & 15.63\cmb & 42.95\cmb & \na & 30.46 & \phantom{0}9.52 & 29.55 & \phantom{0}9.60 & 5.81\cmb & 18.98\cmb & 18.14\cmb \\
SimpleHand~\citep{zhou2024simplehand} & \na & 13.13 & 6.25 & 12.62 & 5.99 & 4.51\cmb & 15.39\cmb & 42.44\cmb & \na & 29.01 & 11.28 & 28.09 & 11.58 & 5.67\cmb & 17.95\cmb & 16.87\cmb \\
\midrule
\textbf{\underline{Object Pose Estimation}} & & & & & & & & & & & & & & & & \\
MegaPose~\citep{labbe2022megapose} & 84.20 & \na & \na & \na & \na & 4.68\cmb & 15.63\cmb & 42.95\cmb & \underline{96.42} & \na & \na & \na & \na & 5.81\cmb & 18.98\cmb & 18.14\cmb \\
GigaPose~\citep{nguyen2024gigapose} & 68.69 & \na & \na & \na & \na & 6.76\cmb & 12.82\cmb & 60.87\cmb & 87.69 & \na & \na & \na & \na & 6.80\cmb & 17.93\cmb & 26.52\cmb \\
FoundPose~\citep{ornek2024foundpose} & 66.24 & \na & \na & \na & \na & 6.58\cmb & 13.87\cmb & 62.64\cmb & 91.25 & \na & \na & \na & \na & 6.42\cmb & 19.37\cmb & 24.53\cmb \\
PoseGAM~\citep{chen2026posegam} & 67.71 & \na & \na & \na & \na & 7.13\cmb & \phantom{0}\textbf{8.57}\cmb & 76.81\cmb & 89.79 & \na & \na & \na & \na & 7.39\cmb & \phantom{0}\underline{8.87}\cmb & 33.52\cmb \\
\midrule
\textbf{\underline{Hand Object Pose Estimation}} & & & & & & & & & & & & & & & & \\
SemiHand~\citep{liu2021semihand} & 82.63 & 13.46 & 5.80 & 13.10 & 5.81 & 5.42 & 16.03 & 42.64 & 68.19 & 29.54 & \phantom{0}9.46 & 28.66 & \phantom{0}9.39 & 6.78 & 17.82 & 89.66 \\
HFL-Net~\citep{lin2023hfl} & 76.02 & 13.27 & 5.74 & 12.88 & 5.72 & 5.90 & 15.39 & 54.19 & 78.43 & 28.45 & \phantom{0}\underline{9.01} & 27.55 & \phantom{0}\underline{8.92} & 6.49 & 20.15 & 55.76 \\
UniHOPE~\citep{wang2025unihope} & 74.20 & 12.92 & 5.79 & 12.55 & 5.77 & 6.00 & 15.31 & 54.92 & 68.79 & 26.23 & \phantom{0}9.60 & 25.40 & \phantom{0}9.45 & 6.48 & 18.02 & 68.87 \\
\textbf{\method\ (ours)} & \underline{92.88} & \textbf{10.12} & \textbf{4.95} & \phantom{0}\textbf{9.80} & \textbf{4.92} & \textbf{3.18} & 14.15 & \phantom{0}\textbf{9.97} & 87.35 & \underline{23.93} & \phantom{0}9.57 & \underline{23.20} & \phantom{0}9.57 & \textbf{4.17} & 19.80 & \textbf{12.76} \\
\textbf{\methodall\ (ours)} & \textbf{95.02} & \underline{10.33} & \underline{5.04} & \underline{10.01} & \underline{5.02} & \underline{3.89} & \underline{11.77} & \underline{11.70} & \textbf{97.31} & \textbf{20.07} & \phantom{0}\textbf{8.93} & \textbf{19.46} & \phantom{0}\textbf{8.85} & 6.87 & \phantom{0}\textbf{6.32} & \underline{15.80} \\
\bottomrule
\end{tabular}}
\end{table*}
\begin{table*}[t]
\centering
\caption{Two-hand comparison on OakInk2 and H2O. Each method is trained on the source dataset and evaluated on the source and target test sets. MRRPE denotes the relative-root error between the hands. \method\ outperforms both bimanual baselines in object pose and relative hand placement, in domain and across datasets.}
\label{tab:hhotransfer}
\par\vspace{-0.5em}
\resizebox{\linewidth}{!}{%
\begin{tabular}{l|ccccccccc|ccccccccc}
\toprule
& \multicolumn{9}{c|}{\textbf{Source}} & \multicolumn{9}{c}{\textbf{Target}} \\
\cmidrule(lr){2-10}\cmidrule(lr){11-19}
& Object & \multicolumn{4}{c}{Hand} & \multicolumn{4}{c|}{Joint} & Object & \multicolumn{4}{c}{Hand} & \multicolumn{4}{c}{Joint} \\
\cmidrule(lr){2-2}\cmidrule(lr){3-6}\cmidrule(lr){7-10}\cmidrule(lr){11-11}\cmidrule(lr){12-15}\cmidrule(lr){16-19}
\textbf{Method} & AUC ($\uparrow$) & MJE ($\downarrow$) & PA-MJE ($\downarrow$) & VPE ($\downarrow$) & PA-VPE ($\downarrow$) & MRRPE ($\downarrow$) & SD ($\downarrow$) & IV ($\downarrow$) & CDev ($\downarrow$) & AUC ($\uparrow$) & MJE ($\downarrow$) & PA-MJE ($\downarrow$) & VPE ($\downarrow$) & PA-VPE ($\downarrow$) & MRRPE ($\downarrow$) & SD ($\downarrow$) & IV ($\downarrow$) & CDev ($\downarrow$) \\
\midrule
\rowcolor{tabgray}\multicolumn{19}{c}{Source: OakInk2~\citep{zhan2024oakink2} $\rightarrow$ Target: H2O~\citep{kwon2021h2o}} \\
\midrule
ArcticNet-SF~\citep{fan2023arctic} & 45.25 & 32.84 & 7.79 & 32.23 & 7.90 & 54.76 & 7.70 & \phantom{0}\underline{9.87} & 49.15 & \phantom{0}3.78 & 46.12 & 11.83 & 44.93 & 11.83 & 131.58 & 4.69 & \underline{14.76} & 26.61 \\
HOPformer~\citep{bansal2026hopformer} & 52.45 & 31.73 & 8.77 & 30.93 & 8.90 & 59.33 & 7.67 & \phantom{0}\textbf{8.50} & 50.85 & \phantom{0}3.08 & 44.43 & 12.06 & 43.58 & 12.25 & 126.77 & \underline{3.05} & 32.54 & 52.32 \\
\textbf{\method\ (ours)} & \underline{65.22} & \underline{14.29} & \underline{6.30} & \underline{13.87} & \underline{6.29} & \underline{43.35} & \underline{6.30} & 15.28 & \underline{34.23} & \underline{29.63} & \underline{21.74} & \phantom{0}\underline{8.68} & \underline{21.42} & \phantom{0}\underline{8.95} & \phantom{0}\underline{84.21} & \textbf{2.97} & 27.25 & \underline{24.64} \\
\textbf{\methodall\ (ours)} & \textbf{87.19} & \textbf{11.18} & \textbf{5.69} & \textbf{10.84} & \textbf{5.69} & \textbf{24.87} & \textbf{5.75} & 13.02 & \textbf{22.14} & \textbf{99.22} & \textbf{10.02} & \phantom{0}\textbf{5.52} & \phantom{0}\textbf{9.95} & \phantom{0}\textbf{5.74} & \phantom{0}\textbf{24.30} & 4.25 & \phantom{0}\textbf{8.27} & \textbf{13.38} \\
\midrule
\rowcolor{tabgray}\multicolumn{19}{c}{Source: H2O~\citep{kwon2021h2o} $\rightarrow$ Target: OakInk2~\citep{zhan2024oakink2}} \\
\midrule
ArcticNet-SF~\citep{fan2023arctic} & 73.82 & 31.08 & 8.74 & 30.14 & 8.71 & 70.06 & \underline{3.72} & 24.53 & 21.59 & 13.41 & 76.72 & 11.95 & 74.37 & 12.05 & 151.29 & \textbf{5.73} & 13.78 & 76.07 \\
HOPformer~\citep{bansal2026hopformer} & 84.06 & 14.49 & \textbf{5.29} & 14.30 & \textbf{5.40} & 54.18 & \textbf{3.70} & \phantom{0}\textbf{7.81} & 17.96 & 10.09 & 32.44 & 10.06 & 31.76 & 10.24 & \phantom{0}88.97 & 8.29 & \phantom{0}\textbf{5.99} & \underline{72.17} \\
\textbf{\method\ (ours)} & \underline{97.95} & \underline{10.61} & 5.74 & \underline{10.52} & 5.94 & \underline{28.03} & 4.35 & \phantom{0}9.79 & \underline{17.02} & \underline{22.56} & \underline{21.85} & \phantom{0}\underline{9.33} & \underline{21.34} & \phantom{0}\underline{9.53} & \phantom{0}\underline{66.29} & 7.66 & 14.44 & 90.25 \\
\textbf{\methodall\ (ours)} & \textbf{99.22} & \textbf{10.02} & \underline{5.52} & \phantom{0}\textbf{9.95} & \underline{5.74} & \textbf{24.30} & 4.25 & \phantom{0}\underline{8.27} & \textbf{13.38} & \textbf{87.19} & \textbf{11.18} & \phantom{0}\textbf{5.69} & \textbf{10.84} & \phantom{0}\textbf{5.69} & \phantom{0}\textbf{24.87} & \underline{5.75} & \underline{13.02} & \textbf{22.14} \\
\bottomrule
\end{tabular}}
\end{table*}
We train \method with frozen encoders in two stages that follow its coarse-to-fine design.
The first stage trains the projections, the shared transformer, and the pose and coarse contact heads.
The second stage then trains the fine head on top of the frozen model, as region distances are informative only once the reconstructed hand and object roughly align.
Freezing the model also keeps the fine head from altering the poses.
Each sample supervises only the hands and object it contains, so one- and two-hand samples from different datasets jointly train a single model.

In the first stage, monocular depth ambiguity mainly affects the absolute placement of the interaction rather than the relative arrangement of the hand and object.
We thus supervise the translation offset from the grasping hand $a$,
\begin{equation}
\mathcal{L}_{\mathrm{rel}}=\rho\big(\Delta\mathbf{t}_o-(\bar{\mathbf{t}}_o-\bar{\mathbf{t}}_a)\big),
\label{eq:rel}
\end{equation}
where $\rho$ is the smooth-$\ell_1$ loss~\citep{girshick2015fast} and bars denote ground truth (GT).
Following prior work, $\mathcal{L}_h$ supervises the MANO parameters and translation~\citep{pavlakos2024hamer}, $\mathcal{L}_o$ the object mesh points posed relative to hand $a$~\citep{xiang2018posecnn, wang2021gdrnet, mo2022es6d}, and $\mathcal{L}_{\mathrm{c}}^h$ per-joint contact with a class-balanced binary cross-entropy (BCE)~\citep{cui2019class, jung2025haco}, as contact labels are imbalanced.
The total loss is
\begin{equation}
\mathcal{L}=\sum_{h}\big(\mathcal{L}_h+\lambda_{\mathrm{c}}\mathcal{L}_{\mathrm{c}}^h\big)+\mathcal{L}_o+\mathcal{L}_{\mathrm{rel}},
\label{eq:loss}
\end{equation}
where $\lambda_{\mathrm{c}}$ weights the contact term.

In the second stage, the fine head is supervised by the GT region contact $\bar{\mathbf{c}}_h\in\{0,1\}^{P}$, where a region is in contact if any of its vertices under the GT poses lies within the contact distance used for the joint labels~\citep{jung2025haco}.
As an auxiliary target, the head also predicts the signed distance $\hat{\mathbf{s}}_h\in\RR^{P}$ of each region to the object, a continuous counterpart of the binary label, and is trained with
\begin{equation}
\mathcal{L}_{\mathrm{fine}}=\sum_{h}\Big(\mathrm{BCE}\big(\tilde{\mathbf{c}}_h,\bar{\mathbf{c}}_h\big)+\rho\big(\hat{\mathbf{s}}_h-\bar{\mathbf{s}}_h\big)\Big),
\label{eq:fineloss}
\end{equation}
where $\bar{\mathbf{s}}_h$ are the minimum signed distances of the regions under the GT poses.
Supp.~\ref{app:setup} provides further training details and the full loss formulations.

\section{Evaluation}
\label{sec:evaluations}

\begin{figure*}[t]
\centering
\includegraphics[width=\textwidth]{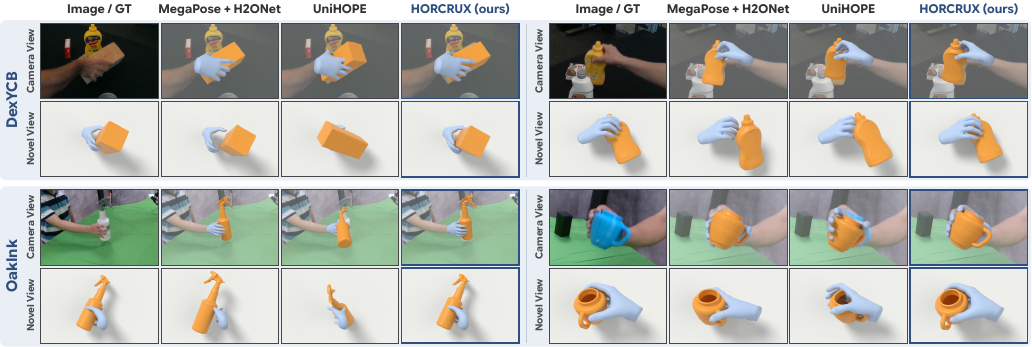}
\par\vspace{-0.5em}
\caption{One-hand qualitative comparison on DexYCB and OakInk against MegaPose~\citep{labbe2022megapose} + H2ONet~\citep{xu2023h2onet} and UniHOPE~\citep{wang2025unihope}. Each prediction is overlaid on the image (camera view) and rendered from a shared rotated viewpoint (novel view). In the novel view, the objects predicted by the baselines drift out of the grasp, whereas \method\ keeps the object in the hand.}
\label{fig:qual_ho}
\end{figure*}
\begin{figure*}[t]
\centering
\includegraphics[width=\textwidth]{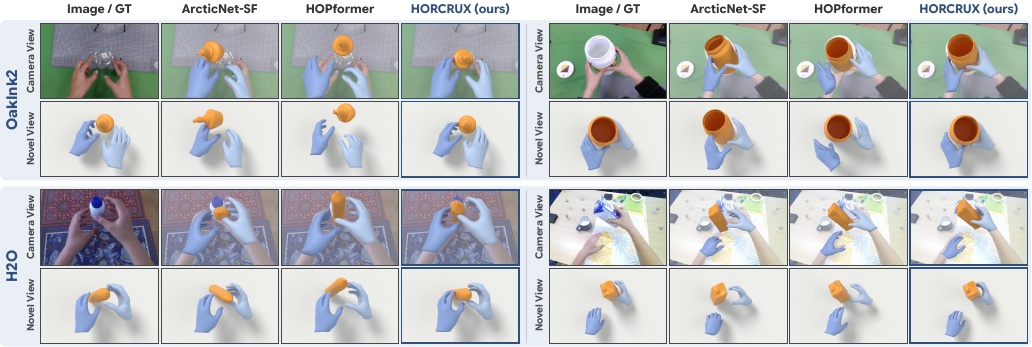}
\par\vspace{-0.5em}
\caption{Two-hand qualitative comparison on OakInk2 and H2O against ArcticNet-SF~\citep{fan2023arctic} and HOPformer~\citep{bansal2026hopformer}, both trained in domain. \method\ keeps the object between the hands, whereas the baselines often misplace it.}
\label{fig:qual_hho}
\end{figure*}

\subsection{Experimental Setup}
\label{subsec:expdetails}

\textbf{Datasets.}\quad
Following~\citet{wang2025unihope}, we evaluate in-domain one-hand interactions on the grasping frames of the DexYCB s0 split~\citep{chao2021dexycb} and HO3Dv2~\citep{hampali2020honnotate}.
To assess generalization, the DexYCB-s0 models are then tested without retraining on HO3Dv2, as well as on OakInk~\citep{yang2022oakink} and the egocentric HOI4D~\citep{liu2022hoi4d}, which contain unseen objects.
For two-hand interactions, we use the official split of H2O~\citep{kwon2021h2o} and a sequence-level split of OakInk2~\citep{zhan2024oakink2}.

\vspace{0.4em}\noindent\textbf{Implementation details.}\quad
\method\ is trained using AdamW~\citep{loshchilov2019adamw} with a cosine-decayed learning rate of $10^{-4}$ and a batch size of 256 on NVIDIA RTX 6000 Ada GPUs.
The foundation encoders~\citep{simeoni2025dinov3,pavlakos2024hamer,chen2026posegam} remain frozen, and hand crops are augmented with random scaling and translation.
Each model is selected by early stopping on its validation split.
\mbox{\methodall} is a single model trained on the union of all training splits, excluding every test frame.
Further details are in Supp.~\ref{app:setup}. The \href{https://shpark97.github.io/Horcrux}{code} will be released upon publication.

\vspace{0.4em}\noindent\textbf{Baselines.}\quad
We compare \method\ against four groups of methods: (i)~hand-pose estimators~\citep{park2022handoccnet, chen2022mobrecon, xu2023h2onet, zhou2024simplehand}; (ii)~zero-shot object-pose estimators given the object mesh~\citep{labbe2022megapose, nguyen2024gigapose, ornek2024foundpose, chen2026posegam}; (iii)~joint hand-object estimators~\citep{liu2021semihand, lin2023hfl, wang2025unihope}; and (iv)~bimanual hand-object estimators~\citep{fan2023arctic, bansal2026hopformer}, which we retrain on each two-hand dataset.
Hand-pose estimators are paired with \mbox{MegaPose}~\citep{labbe2022megapose} and object-pose estimators with H2ONet~\citep{xu2023h2onet}, the strongest counterparts, and marked with~$\dagger$.

\vspace{0.4em}\noindent\textbf{Evaluation protocol.}\quad
For the hand, we report mean joint and vertex position errors (MJE, VPE) and their Procrustes-aligned variants (PA-MJE, PA-VPE).
For the object, we report the AUC of ADD-0.5D~\citep{wang2025unihope}, the percentage of frames whose average point distance (ADD) lies within half the object diameter.
For the joint pose, simulation displacement (SD) is the distance the object moves under simulated gravity with the hand fixed, and intersection volume (IV) is the volume shared by the hand and object meshes~\citep{hasson2019obman}.
Contact deviation (CDev)~\citep{fan2023arctic} averages the predicted hand-object distance over the GT contact vertices.
The full protocol and metric definitions are in Supp.~\ref{app:setup}.

\subsection{Hand-Object Reconstruction}
\label{subsec:mainresults}
\label{subsec:indomain}
\textbf{One-hand interaction.}\quad
We first evaluate in-domain one-hand reconstruction on DexYCB and HO3D.
As shown in \mbox{Table~\ref{tab:indomainpose}}, \mbox{\method} outperforms every baseline on DexYCB in all hand and object metrics, with an MJE of 10.12 and a \mbox{PA-MJE} of 4.95.
Its object AUC of 92.88 is 10.25 above the strongest joint estimator SemiHand~\citep{liu2021semihand} and 8.68 above even MegaPose~\citep{labbe2022megapose}, which estimates only the object from the same mesh.
The margin widens in the joint metrics, where its CDev of 9.97 is less than a quarter of the lowest baseline CDev of 42.44.
Even the joint estimators, which predict the hand and object together, fail to keep the object in the grasp.
PoseGAM~\citep{chen2026posegam} attains the lowest IV only by leaving the object far from the hand, with the largest CDev of 76.81.
On HO3D, \method\ again leads every baseline with an MJE of 23.93 and a VPE of 23.20 at a comparable \mbox{PA-MJE} of 9.57.
Its object AUC of 87.35 trails the zero-shot object-pose estimators, which reach up to 96.42, but is 8.92 above the strongest joint estimator HFL-Net~\citep{lin2023hfl}.
Its CDev and SD are also the lowest.
Trained on all datasets, \mbox{\methodall} attains the highest object AUC of 97.31 and the lowest PA-MJE of 8.93 among all methods, lowering the MJE by a further 3.86.
These gains stem from reasoning jointly over the hand and the frozen foundation features of the object mesh, and from posing the object relative to the hand, which keeps it in the grasp.

\vspace{0.4em}\noindent\textbf{Two-hand interaction.}\quad
We then evaluate in-domain two-hand reconstruction on OakInk2 and H2O and report the relative-root error between the hands (MRRPE)~\citep{fan2023arctic}.
As shown in Table~\ref{tab:hhotransfer}, \method\ attains an object AUC of 65.22 on OakInk2 and 97.95 on H2O, outperforming the stronger baseline HOPformer~\citep{bansal2026hopformer} by 12.77 and 13.89.
It also more than halves the MJE on OakInk2 to 14.29 and lowers the MRRPE to 43.35 and 28.03.
These gains need no dedicated bimanual module, since the shared transformer relates both hands and the object as it does for one hand.
On H2O, its \mbox{PA-MJE} and \mbox{PA-VPE} are on par with \mbox{HOPformer}, within 0.54, while its MJE is 3.88 lower.
In Figures~\ref{fig:qual_ho} and~\ref{fig:qual_hho}, all methods overlay the object well in the camera view, but only \method\ keeps it in the grasp from a novel view.
Additional qualitative results are in Supp.~\ref{app:more_qual}.

\setcounter{topnumber}{2}
\suppressfloats[t]
\begin{table}[t]
\centering
\caption{Coarse contact on DexYCB and OakInk. We report F1 over the 21 joints under the 1\,cm labeling rule of HACO~\citep{jung2025haco} (F1@1cm) and the stricter 3\,mm rule of CDev~\citep{fan2023arctic} (F1@3mm). The joint estimators are scored on the contact implied by their meshes.}
\label{tab:coarsecontact}
\par\vspace{-0.5em}
\resizebox{\columnwidth}{!}{%
\begin{tabular}{l|cc|cc}
\toprule
& \multicolumn{2}{c|}{\textbf{DexYCB}~\citep{chao2021dexycb}} & \multicolumn{2}{c}{\textbf{OakInk}~\citep{yang2022oakink}} \\
\cmidrule(lr){2-3}\cmidrule(lr){4-5}
\textbf{Method} & F1@1cm ($\uparrow$) & F1@3mm ($\uparrow$) & F1@1cm ($\uparrow$) & F1@3mm ($\uparrow$) \\
\midrule
SemiHand~\citep{liu2021semihand} & 63.06 & 46.10 & 29.10 & 18.83 \\
HFL-Net~\citep{lin2023hfl} & 57.96 & 41.67 & 29.11 & 18.90 \\
UniHOPE~\citep{wang2025unihope} & 55.95 & 40.20 & 27.85 & 18.34 \\
HACO~\citep{jung2025haco} & \underline{84.56} & \underline{68.09} & \underline{77.62} & \underline{59.15} \\
\method\ (ours) & \textbf{85.06} & \textbf{74.97} & \textbf{85.67} & \textbf{73.62} \\
\bottomrule
\end{tabular}}
\end{table}
\begin{figure}[t]
\centering
\begin{subfigure}[t]{0.49\linewidth}
\centering
\includegraphics[width=\linewidth]{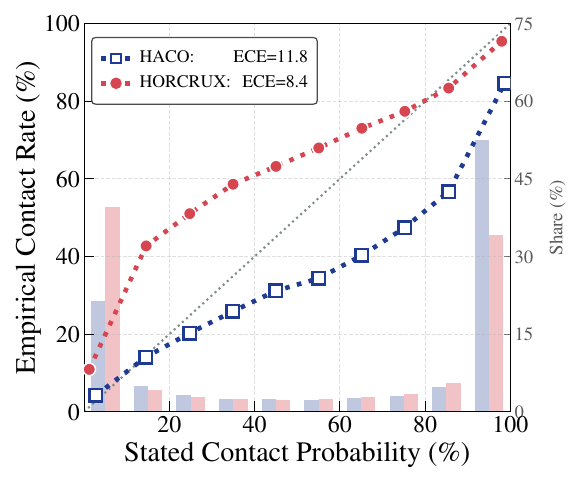}
\caption{DexYCB}
\label{fig:contact_rel_dexycb}
\end{subfigure}\hfill
\begin{subfigure}[t]{0.49\linewidth}
\centering
\includegraphics[width=\linewidth]{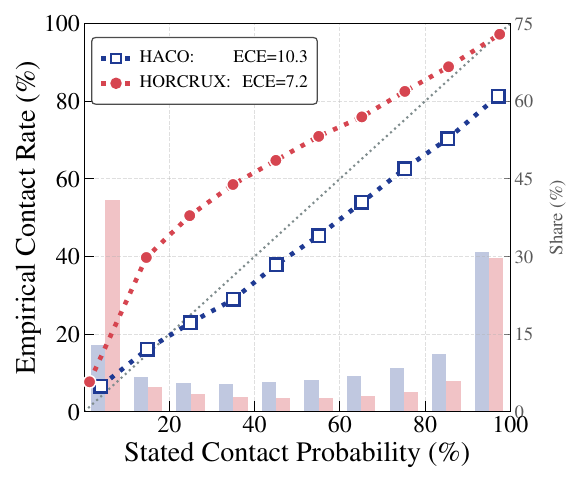}
\caption{OakInk}
\label{fig:contact_rel_oakink}
\end{subfigure}
\par\vspace{-0.5em}
\caption{Calibration of the coarse contact of \method\ and HACO~\citep{jung2025haco} under the 1\,cm rule. Each point is the empirical contact rate of a probability bin, and each bar is its share of predictions. The expected calibration error (ECE) in the legend weights each bin by this share and is lower for \method\ on both datasets.}
\label{fig:contact_analysis}
\end{figure}
\begin{table}[t]
\centering
\caption{Ablation of coarse-to-fine contact estimation. We report F1 over the finer hand regions under the rules of Table~\ref{tab:coarsecontact}. Combining the coarse contact with the region distances of the reconstruction (\textit{+ geometry}) gives the highest F1 in three of the four settings.}
\label{tab:finecontact}
\par\vspace{-0.5em}
\resizebox{\columnwidth}{!}{%
\begin{tabular}{l|cc|cc}
\toprule
& \multicolumn{2}{c|}{\textbf{DexYCB}~\citep{chao2021dexycb}} & \multicolumn{2}{c}{\textbf{OakInk}~\citep{yang2022oakink}} \\
\cmidrule(lr){2-3}\cmidrule(lr){4-5}
\textbf{Method} & F1@1cm ($\uparrow$) & F1@3mm ($\uparrow$) & F1@1cm ($\uparrow$) & F1@3mm ($\uparrow$) \\
\midrule
Coarse only & 78.97 & 54.92 & 77.49 & 49.21 \\
Fine only & \textbf{80.75} & \underline{60.98} & 80.65 & 56.13 \\
Coarse-to-fine & 80.37 & 60.44 & \underline{81.55} & \underline{56.48} \\
\textit{+ geometry (ours)} & \underline{80.59} & \textbf{61.63} & \textbf{82.32} & \textbf{57.58} \\
\bottomrule
\end{tabular}}
\end{table}
\begin{figure}[t]
\centering
\includegraphics[width=\columnwidth]{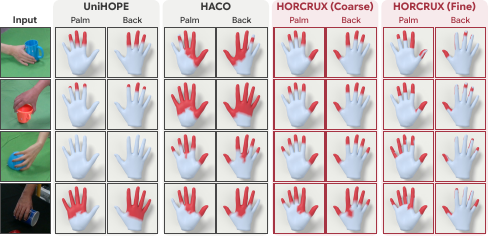}
\par\vspace{-0.5em}
\caption{Contact on DexYCB and OakInk, shown in \textcolor{red}{red} on a canonical hand viewed from the palm and the back. The contact implied by the meshes of UniHOPE~\citep{wang2025unihope} misses much of the grasp, and the contact of HACO~\citep{jung2025haco} spreads over the hand, whereas the fine contact of \method\ keeps only the touching palmar regions.}
\label{fig:contactqual}
\end{figure}

\subsection{Contact Estimation}
\label{subsec:contactacc}
\textbf{Coarse contact.}\quad
Beyond the poses, \method\ predicts which hand joints touch the object.
We compare this coarse contact with the contact implied by the meshes of the joint estimators and with the dedicated estimator HACO~\citep{jung2025haco}.
All are scored under the 1\,cm labeling rule of HACO and the stricter 3\,mm rule of CDev~\citep{fan2023arctic}.
As shown in Table~\ref{tab:coarsecontact}, \method\ attains the highest F1 under both rules on both datasets.
On DexYCB, it is on par with HACO under the 1\,cm rule but outperforms it by 6.88 under the 3\,mm rule, and on OakInk it leads by 8.05 and 14.47 under the two rules.
Its advantage thus lies in localizing the contact precisely.
The mesh-implied contact of the joint estimators reaches at most 63.06 on DexYCB, consistent with their CDev of at least 42.64.
Such contact inherits the pose errors of the meshes, whereas \method\ predicts contact directly from its latent tokens.
Moreover, Figure~\ref{fig:contact_analysis} shows that its contact probabilities are better calibrated than those of HACO, lowering the expected calibration error by 3.4 on DexYCB and 3.1 on OakInk.
This makes them a reliable prior for the fine head.

\begin{table*}[t]
\centering
\caption{Cross-dataset transfer of the DexYCB-s0 models to OakInk, HO3D, and HOI4D without retraining. \method\ outperforms the joint estimators in object pose, hand pose, and contact deviation on all three datasets. The object-pose margin is small on HO3D, whose objects also appear in DexYCB, and widens on the unseen objects of OakInk and HOI4D.}
\label{tab:domaingen}
\par\vspace{-0.5em}
\resizebox{\linewidth}{!}{%
\begin{tabular}{l|cccccc|cccccc|cccccc}
\toprule
& \multicolumn{6}{c|}{\textbf{OakInk}~\citep{yang2022oakink}} & \multicolumn{6}{c|}{\textbf{HO3D}~\citep{hampali2020honnotate}} & \multicolumn{6}{c}{\textbf{HOI4D}~\citep{liu2022hoi4d}} \\
\cmidrule(lr){2-7}\cmidrule(lr){8-13}\cmidrule(lr){14-19}
& Object Pose & \multicolumn{2}{c}{Hand Pose} & \multicolumn{3}{c|}{Joint Pose} & Object Pose & \multicolumn{2}{c}{Hand Pose} & \multicolumn{3}{c|}{Joint Pose} & Object Pose & \multicolumn{2}{c}{Hand Pose} & \multicolumn{3}{c}{Joint Pose} \\
\cmidrule(lr){2-2}\cmidrule(lr){3-4}\cmidrule(lr){5-7}\cmidrule(lr){8-8}\cmidrule(lr){9-10}\cmidrule(lr){11-13}\cmidrule(lr){14-14}\cmidrule(lr){15-16}\cmidrule(lr){17-19}
\textbf{Method} & AUC ($\uparrow$) & PA-MJE ($\downarrow$) & PA-VPE ($\downarrow$) & SD ($\downarrow$) & IV ($\downarrow$) & CDev ($\downarrow$) & AUC ($\uparrow$) & PA-MJE ($\downarrow$) & PA-VPE ($\downarrow$) & SD ($\downarrow$) & IV ($\downarrow$) & CDev ($\downarrow$) & AUC ($\uparrow$) & PA-MJE ($\downarrow$) & PA-VPE ($\downarrow$) & SD ($\downarrow$) & IV ($\downarrow$) & CDev ($\downarrow$) \\
\midrule
SemiHand~\citep{liu2021semihand} & 40.50 & 7.97 & 8.34 & 8.85 & 10.82 & 136.21 & 86.81 & 9.64 & 9.54 & 7.18 & 23.09 & 44.82 & 15.57 & 10.60 & 10.95 & 9.63 & \phantom{0}5.29 & 214.96 \\
HFL-Net~\citep{lin2023hfl} & 37.83 & 8.18 & 8.53 & 9.01 & \phantom{0}\underline{8.94} & 156.37 & 81.16 & 9.86 & 9.81 & 7.17 & 20.47 & 48.29 & 13.52 & 11.57 & 11.83 & 9.81 & \phantom{0}\underline{2.85} & 234.34 \\
UniHOPE~\citep{wang2025unihope} & 38.81 & 7.96 & 8.32 & 9.09 & \phantom{0}\textbf{8.86} & 158.83 & 79.57 & 9.81 & 9.72 & 7.80 & 18.90 & 53.68 & 14.17 & 10.72 & 10.99 & 9.80 & \phantom{0}\textbf{2.76} & 245.02 \\
\textbf{\method\ (ours)} & \underline{93.29} & \underline{7.16} & \underline{7.61} & \underline{5.16} & 16.21 & \phantom{0}\underline{17.55} & \underline{88.75} & \textbf{8.46} & \textbf{8.37} & \textbf{5.62} & \underline{17.09} & \textbf{13.71} & \underline{67.68} & \phantom{0}\underline{9.16} & \phantom{0}\underline{9.63} & \textbf{6.27} & 17.95 & \phantom{0}\underline{17.45} \\
\textbf{\methodall\ (ours)} & \textbf{99.97} & \textbf{4.81} & \textbf{4.85} & \textbf{4.50} & 11.43 & \phantom{0}\textbf{10.21} & \textbf{97.31} & \underline{8.93} & \underline{8.85} & \underline{6.87} & \phantom{0}\textbf{6.32} & \underline{15.80} & \textbf{97.36} & \phantom{0}\textbf{6.32} & \phantom{0}\textbf{6.35} & \underline{6.56} & \phantom{0}4.36 & \phantom{0}\textbf{10.62} \\
\bottomrule
\end{tabular}}
\end{table*}
\vspace{0.4em}\noindent\textbf{Fine contact.}\quad
Defined per joint, the coarse contact cannot determine whether a joint touches the object on its palmar or dorsal side.
The fine head resolves it over finer hand regions using the region distances of the reconstruction, and Table~\ref{tab:finecontact} compares this design with three variants.
\emph{Coarse only} broadcasts the per-joint contact to the regions, \emph{Fine only} predicts the region contact from latent tokens, and \emph{Coarse-to-fine} resolves the coarse contact without the region distances.
The full model attains the highest F1 in three of the four settings and falls 0.16 short of \emph{Fine only} on DexYCB under the 1\,cm rule.
Without the region distances, \emph{Coarse-to-fine} is lower in all four settings, and \emph{Coarse only} drops by up to 8.37 under the 3\,mm rule.
Both cues are thus needed, as the coarse contact identifies the touching joints and the reconstructed geometry localizes it on the hand surface.
Figure~\ref{fig:contactqual} shows how the two cues combine.
The coarse contact covers the touching joints on both the palm and the back, and the fine contact clears most of the back while keeping the touching palmar regions.
In contrast, HACO spreads contact over the back, and the mesh-implied contact of UniHOPE misses much of the grasp.
Each relies on a single cue, whereas \method\ combines both to localize the contact precisely.

\begin{figure}[t]
\centering
\includegraphics[width=\columnwidth]{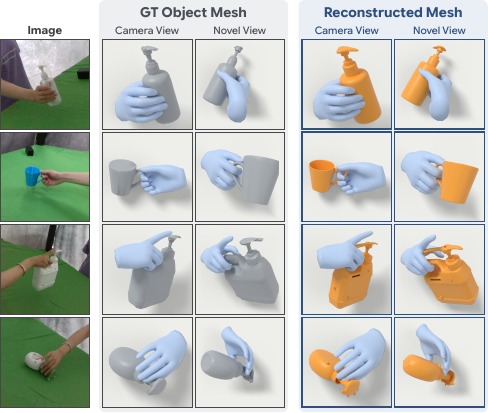}
\par\vspace{-0.5em}
\caption{Approximate object meshes on OakInk test frames. \mbox{\method} takes the GT object mesh (\textcolor{meshgrey}{grey}) or the mesh that \mbox{SAM 3D}~\citep{sam3d2025} reconstructs from the photograph (\textcolor{meshorange}{\mbox{orange}}). Although the two meshes differ in shape, the predicted hand and object poses nearly coincide in the camera and novel views.}
\label{fig:weakwild}
\end{figure}
\setcounter{topnumber}{99}

\subsection{Generalization to Unseen Objects}
\label{subsec:domaingen}
\label{subsec:hhotransfer}
\textbf{Cross-dataset transfer.}\quad
We apply each DexYCB-s0 model to OakInk, HO3D, and HOI4D without retraining and swap the source and target datasets of the two-hand models.
As shown in Table~\ref{tab:domaingen}, the strongest joint estimator \mbox{SemiHand}~\citep{liu2021semihand} is within 1.94 AUC of \method\ on HO3D, whose objects also appear in DexYCB.
On the unseen objects of OakInk and HOI4D, \method\ reaches an object AUC of 93.29 and 67.68 with a lower hand error, exceeding SemiHand by 52.79 and 52.11.
The lower IV of the baselines there stems from objects left far from the hand, as their CDev of at least 136.21 shows.
Across the two-hand datasets in Table~\ref{tab:hhotransfer}, the bimanual baselines likewise reach at most 13.41 AUC, whereas \method\ attains 22.56 and 29.63.
The margin thus grows with object novelty, as the baselines rely on the appearance of their training objects.
\method\ instead keeps the object in the grasp by posing it relative to the hand and recovers its rotation from the geometry encoded by a frozen object-pose foundation model.
Retrained without this encoding, the model still keeps the object in the grasp, but its rotation error rises sharply (see Supp.~\ref{app:analyses}).

\vspace{0.4em}\noindent\textbf{Approximate object meshes.}\quad
\label{subsec:wild}
Since \method\ encodes the object through its geometry, the same model can take an approximate mesh of the held object in place of a scanned one.
We test this on 500 OakInk test frames, replacing the GT object mesh with one reconstructed from the photograph by a state-of-the-art image-to-3D model, \mbox{SAM 3D}~\citep{sam3d2025} or \mbox{Hunyuan3D 2.1}~\citep{hunyuan3d2025v21}.
Because ADD is undefined once the input mesh differs from the GT, we measure the object \mbox{error} by the Chamfer distance (CD) to the posed GT object, which covers both the pose error and the shape difference.
As shown in Table~\ref{tab:weakwildq}, the reconstructed meshes change the \mbox{PA-MJE} and \mbox{PA-VPE} by at most 0.07, whereas the CD rises from 19.01 with the GT mesh to 27.04 with \mbox{SAM 3D} and 45.20 with \mbox{Hunyuan3D 2.1}.
With the smaller rise of 8.03, the \mbox{SAM 3D} mesh is the more practical substitute for a scan.
Figure~\ref{fig:weakwild} shows that the hand and object poses predicted with this mesh closely match those obtained with the GT mesh in both the camera and novel views, even though the two meshes differ in shape.%
\begin{table}[t]
\centering
\caption{Accuracy of \method\ on 500 OakInk test frames with the GT object mesh or one reconstructed from the photograph. CD and Center are the Chamfer and centroid distances to the posed GT object. Reconstructed meshes leave the hand error nearly unchanged but raise the object error.}
\label{tab:weakwildq}
\par\vspace{-0.5em}
\resizebox{\columnwidth}{!}{%
\begin{tabular}{l|cc|cc}
\toprule
& \multicolumn{2}{c|}{\textbf{Hand Pose}} & \multicolumn{2}{c}{\textbf{Object Pose}} \\
\cmidrule(lr){2-3}\cmidrule(lr){4-5}
\textbf{Object Model} & PA-MJE ($\downarrow$) & PA-VPE ($\downarrow$) & CD ($\downarrow$) & Center ($\downarrow$) \\
\midrule
GT Object Mesh & \textbf{5.68} & \textbf{5.76} & \textbf{19.01} & \textbf{36.72} \\
SAM 3D~\citep{sam3d2025} & 5.71 & 5.79 & 27.04 & 49.16 \\
Hunyuan3D 2.1~\citep{hunyuan3d2025v21} & 5.75 & 5.81 & 45.20 & 66.45 \\
\bottomrule
\end{tabular}}
\end{table}

\section{Conclusion}
\label{sec:conclusion}
\textbf{Further analyses.}\quad
Beyond the main experiments, the supplementary material provides additional analyses of \mbox{\method}.
First, we evaluate robustness to occlusion, grouping the test frames by the fraction of the hand or object silhouette hidden by the other.
Second, we report the object pose accuracy under stricter thresholds.
Third, we ablate the frozen hand and object features by removing each and retraining the model to isolate its contribution.
Finally, we provide further qualitative results on one-hand and two-hand interactions, contact estimation, and approximate object meshes.
Comprehensive results are available in Supp.~\ref{app:more_qual} and~\ref{app:analyses}.

\vspace{0.4em}\noindent\textbf{Closing remarks.}\quad
We presented \method, a unified feed-forward transformer that reconstructs one or two hands with an unseen object and localizes their contact precisely, which existing methods achieve only in part.
Joint reasoning over frozen foundation features encodes the object from its geometry rather than object-specific knowledge and keeps it aligned with the hands.
The model thus extends to unseen objects, even with image-reconstructed meshes.
Coarse-to-fine contact estimation then draws on this reconstruction rather than the image alone, resolving per-joint contact from latent features over finer hand regions to localize it on the hand surface.
Across seven benchmarks, \method\ outperforms prior joint estimators in object and hand pose, and its contact is more accurate than that of a dedicated estimator.
These findings suggest that foundation features let hand-object reconstruction generalize to unseen objects, and that this reconstruction in turn localizes contact precisely.

\section*{LLM Usage Statement}
We used Claude (Anthropic) to assist with coding, running experiments, creating figures, improving readability, and formatting references.
The authors directed the research, reviewed all AI-assisted work, and verified the code and results.
They take full responsibility for the final content, including its text, claims, and artifacts.

{
    \small
    \bibliographystyle{ieeenat_fullname}
    \bibliography{main}
}

\appendix
\newsavebox\fitwbox
\newcommand{\fitw}[1]{\sbox\fitwbox{#1}\ifdim\wd\fitwbox>\linewidth\resizebox{\linewidth}{!}{\usebox\fitwbox}\else\usebox\fitwbox\fi}
\newcommand{\appfig}[1]{\par\vspace{0.6em}\noindent\begin{minipage}{\textwidth}\captionsetup{type=figure}\centering #1\end{minipage}\par}
\clearpage
\onecolumn
\begin{center}
    {\Large \bf Generalizable Feed-Forward 3D Hand-Object Reconstruction\\ with Coarse-to-Fine Contact Estimation} \\
    \vspace{0.8em}
    {\large Supplementary Material} \\
    \vspace{0.6em}
\end{center}
\raggedbottom

\section{Additional Qualitative Results}
\label{app:more_qual}
This section provides additional qualitative results of \method.
We first compare \method\ with the one-hand and bimanual baselines in Supp.~\ref{app:more_qual_pose} and~\ref{app:more_qual_hho}.
We then visualize its coarse and fine contact alongside the contact baselines in Supp.~\ref{app:more_qual_contact}.
Finally, we show reconstructions with approximate object meshes in Supp.~\ref{app:more_weakwild} and on internet photographs in Supp.~\ref{sec:wild}.

\subsection{One-Hand Interaction}
\label{app:more_qual_pose}
Figures~\ref{fig:app_qual_1} and~\ref{fig:app_qual_2} compare \method\ with MegaPose~\citep{labbe2022megapose} + H2ONet~\citep{xu2023h2onet} and UniHOPE~\citep{wang2025unihope} on DexYCB and OakInk, respectively.
The baselines often let the object drift out of the grasp or misestimate its rotation, which the novel view reveals most clearly.
In contrast, \method\ keeps the object in the hand and stays close to the GT in both views.

\vspace*{\fill}
\appfig{
\includegraphics[width=\textwidth]{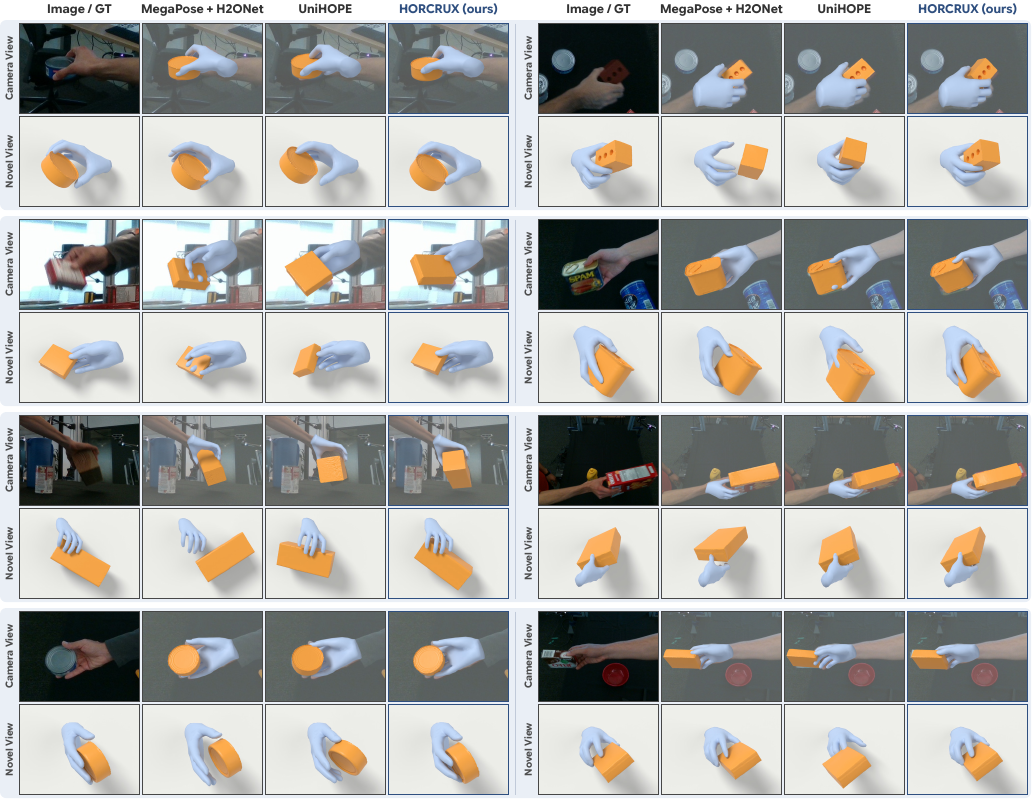}
\par\vspace{-0.5em}
\caption{We compare \method\ with the one-hand baselines on DexYCB~\citep{chao2021dexycb}. Each prediction is overlaid on the image in the camera view (top) and rendered from a shared novel view (bottom).}
\label{fig:app_qual_1}}
\vfill
\clearpage

\vspace*{\fill}
\appfig{
\includegraphics[width=\textwidth]{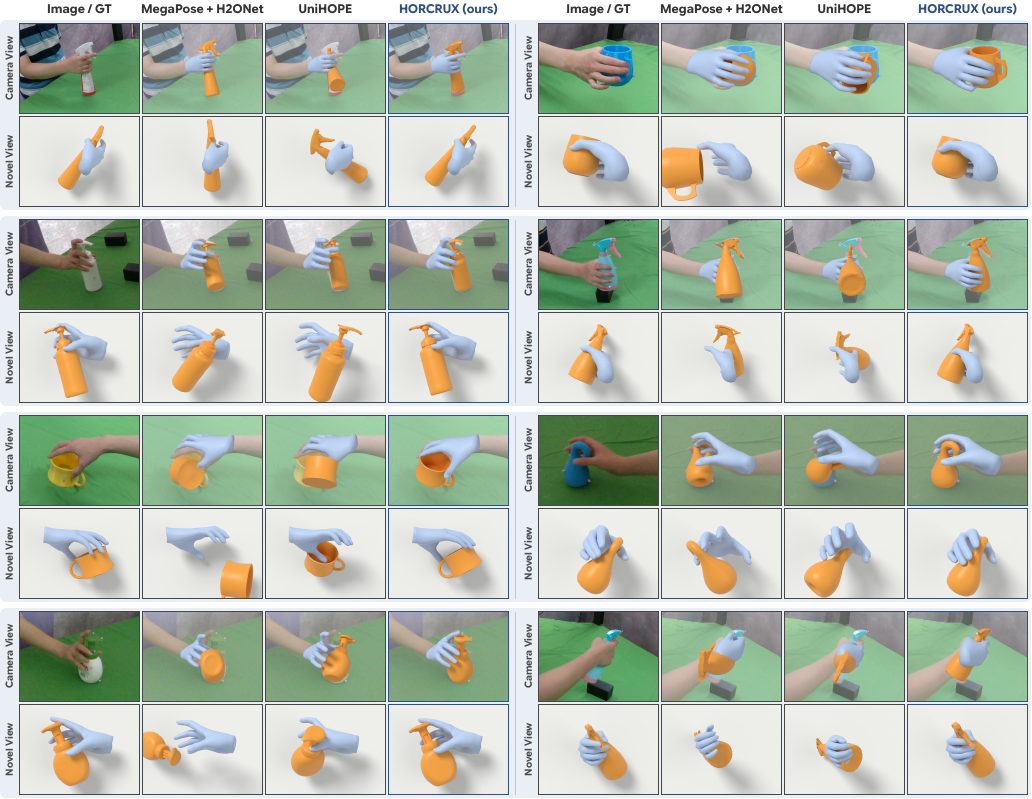}
\par\vspace{-0.5em}
\caption{We compare \method\ with the one-hand baselines on OakInk~\citep{yang2022oakink}.}
\label{fig:app_qual_2}}
\vfill
\clearpage

\subsection{Two-Hand Interaction}
\label{app:more_qual_hho}
Figure~\ref{fig:app_qual_hho} compares \method\ with ArcticNet-SF~\citep{fan2023arctic} and HOPformer~\citep{bansal2026hopformer} on OakInk2 and H2O.
As in the one-hand case, the baselines often misplace the object, whereas \method\ keeps it between the hands.

\vspace*{\fill}
\appfig{
\includegraphics[width=\textwidth]{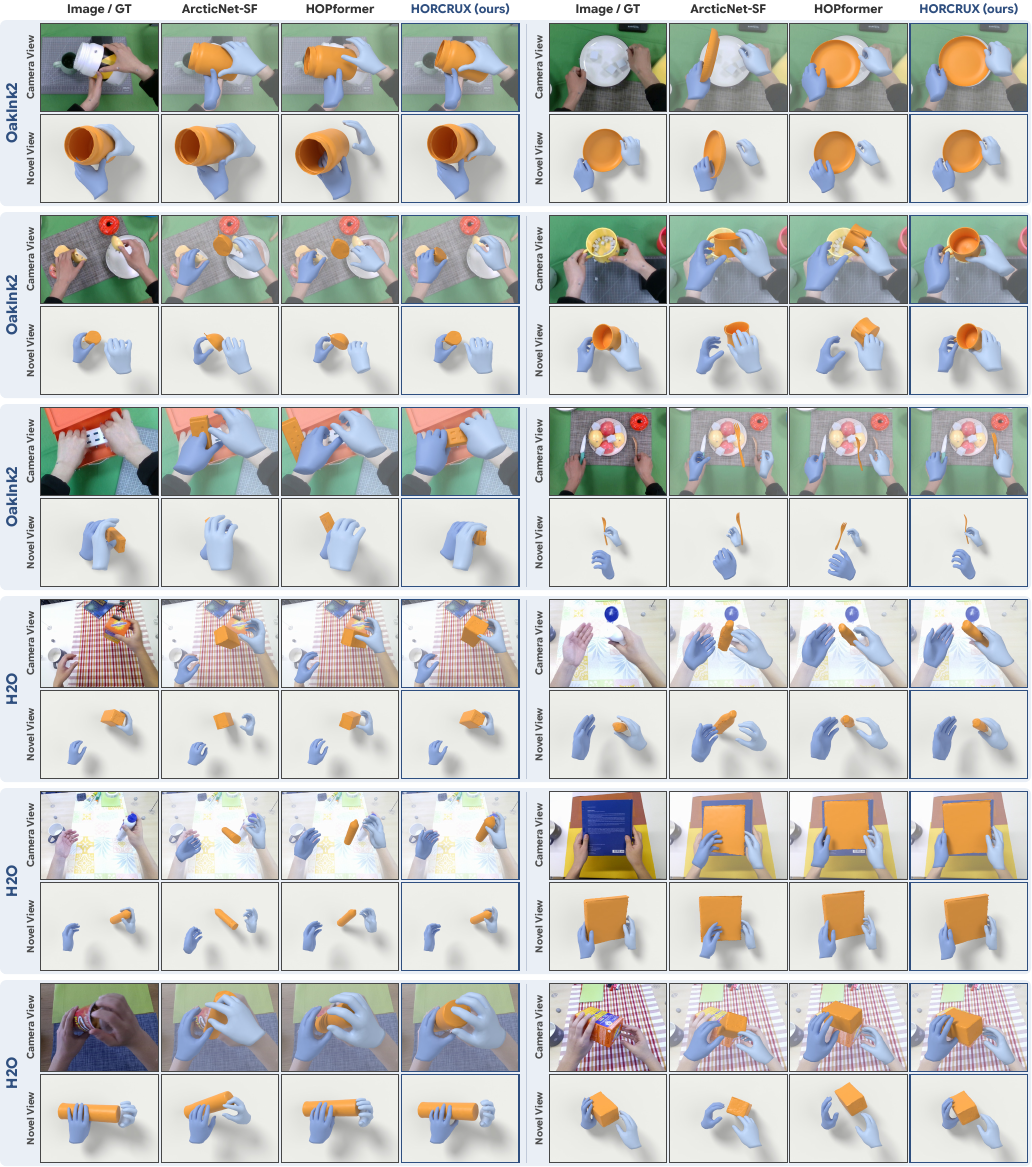}
\par\vspace{-0.5em}
\caption{We compare \method\ with the bimanual baselines on OakInk2~\citep{zhan2024oakink2} and H2O~\citep{kwon2021h2o}.}
\label{fig:app_qual_hho}}
\vfill
\clearpage

\subsection{Contact Estimation}
\label{app:more_qual_contact}
Figure~\ref{fig:app_contact} compares the coarse and fine contact of \method\ with HACO~\citep{jung2025haco} and the mesh-implied contact of UniHOPE~\citep{wang2025unihope} on DexYCB and OakInk.
The coarse contact identifies the touching joints but, like the baselines, often spreads over whole finger segments on both the palm and the back.
The fine contact refines it on the hand surface, clearing most of the back and pinpointing the finger pads and palm regions that touch the object.

\vspace*{\fill}
\appfig{
\includegraphics[width=\textwidth]{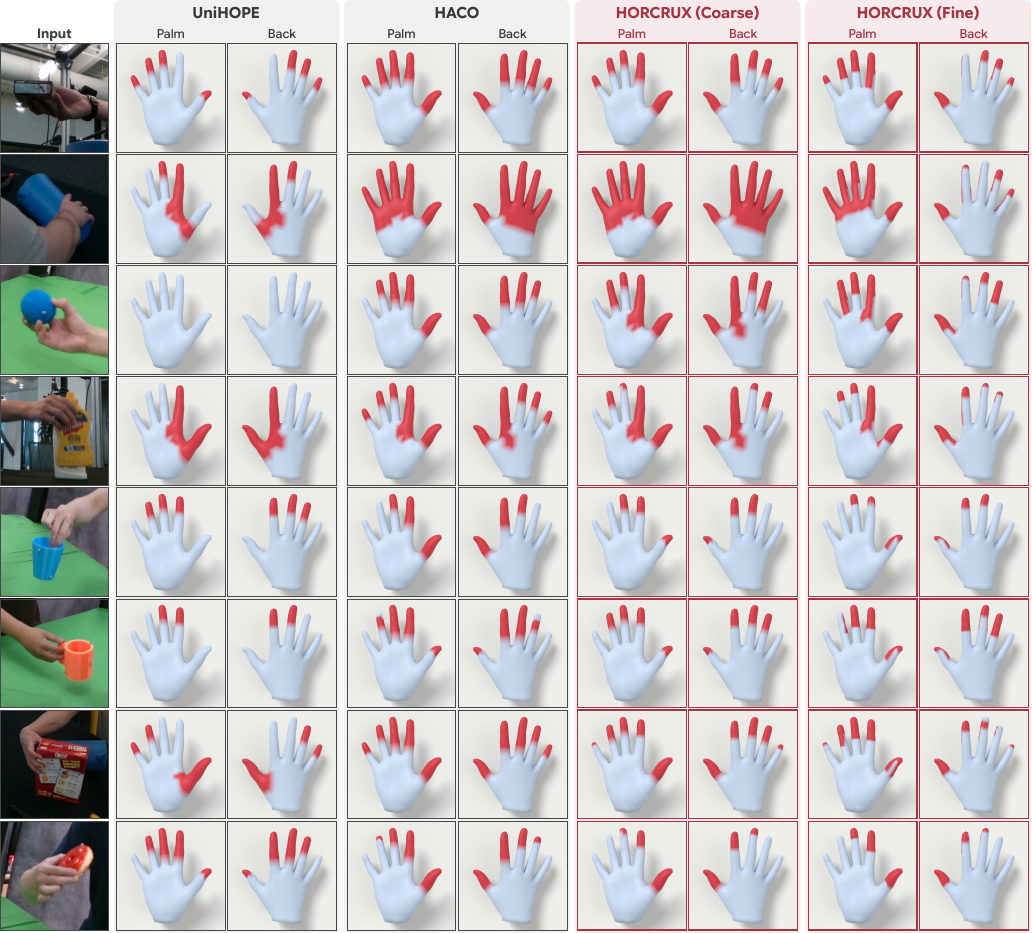}
\par\vspace{-0.5em}
\caption{We compare the coarse and fine contact of \method\ with the contact baselines on DexYCB~\citep{chao2021dexycb} and OakInk~\citep{yang2022oakink}, shown in \textcolor{contactred}{red} on a canonical hand viewed from the palm and the back.}
\label{fig:app_contact}}
\vfill
\clearpage

\subsection{Approximate Object Meshes}
\label{app:more_weakwild}
Figure~\ref{fig:app_weakwild_oakink} extends Figure~\ref{fig:weakwild} to a wider range of OakInk objects (\eg, mugs, bottles, binoculars, a wrench, and a hammer).
Although the reconstructed meshes differ from the GT in shape and detail, \method\ predicts nearly the same hand and object poses with either mesh in both views.

\vspace*{\fill}
\appfig{
\includegraphics[width=\textwidth]{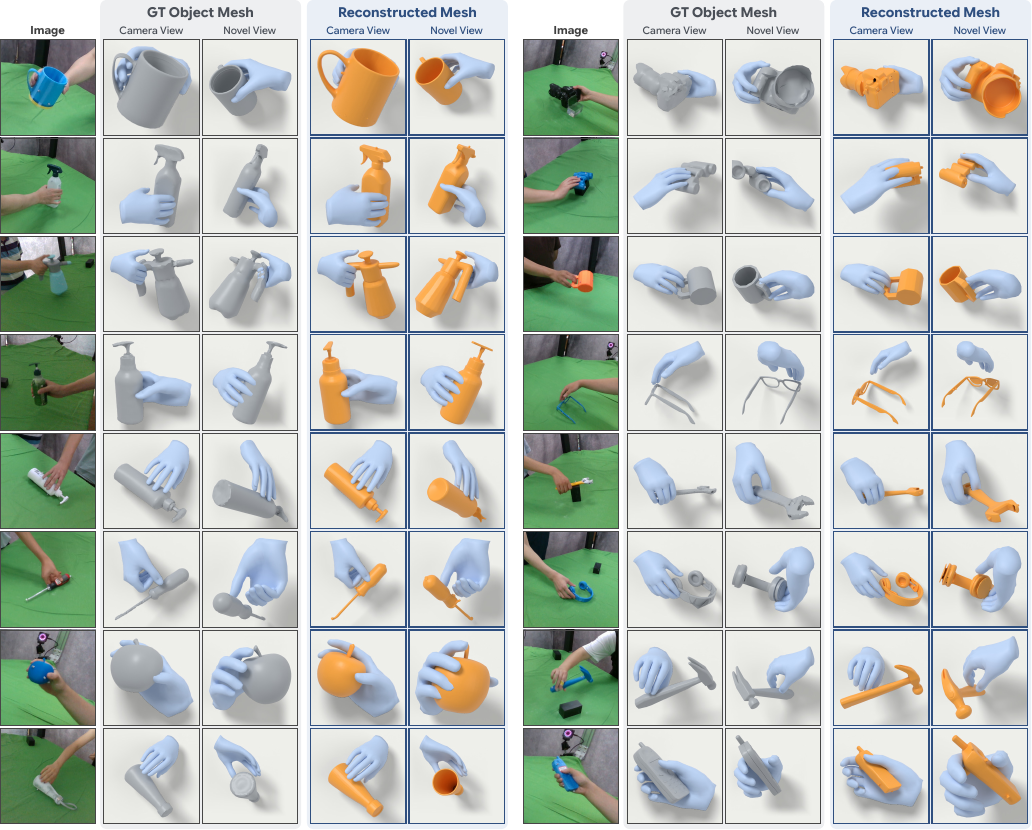}
\par\vspace{-0.5em}
\caption{We apply \method\ to diverse OakInk~\citep{yang2022oakink} objects, using either the GT object mesh (\textcolor{meshgrey}{grey}) or the mesh that \mbox{SAM 3D}~\citep{sam3d2025} reconstructs from the input image (\textcolor{meshorange}{orange}).}
\label{fig:app_weakwild_oakink}}
\vfill
\clearpage
\subsection{Internet Photographs}
\label{sec:wild}
Figure~\ref{fig:wild} applies \method\ to internet photographs, where \mbox{SAM 3D}~\citep{sam3d2025} reconstructs the held object from each photograph.
These results are qualitative only, as the photographs provide neither GT hand and object poses nor camera intrinsics.
Because these photographs differ from the training datasets, the fine finger articulation can be imprecise.
\method\ nevertheless recovers a rough but plausible interaction, with the hand placed around the held object in an approximate grasp.
\appfig{
\includegraphics[width=0.62\textwidth]{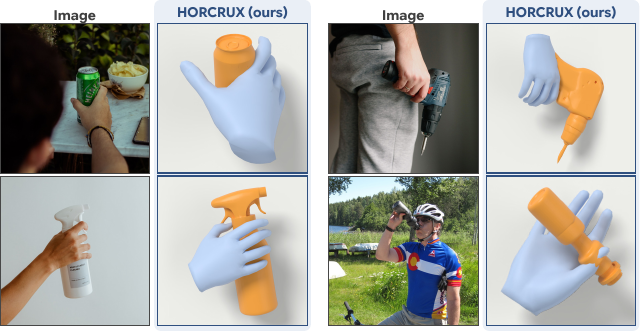}
\par\vspace{-0.5em}
\caption{We apply \method\ to internet photographs, using the object mesh that \mbox{SAM 3D}~\citep{sam3d2025} reconstructs from each photograph.}
\label{fig:wild}}
\section{Experimental Details}
\label{app:setup}
This section extends the experimental setup of the main paper with additional details of \method.
We first describe the implementation details in Supp.~\ref{app:impl}.
We then present the loss terms in Supp.~\ref{app:lossforms}.
Next, we summarize the datasets with their splits and training filters in Supp.~\ref{app:data}.
Finally, we detail the evaluation protocol and metrics in Supp.~\ref{subsec:protocol}.

\subsection{Implementation Details}
\label{app:impl}
\textbf{Architecture.}\quad
The DINOv3 encoder $E_{\mathrm{img}}$ is a frozen ViT-B/16~\citep{simeoni2025dinov3} applied to $I$ at $224\times224$, and its scene tokens serve only as context for the other tokens.
The hand encoder $E_h$ of Eqn.~\eqref{eq:fuse} combines the same DINOv3 with a frozen hand-pose backbone~\citep{pavlakos2024hamer} on the crop $I_h$ at $256\times256$, which also covers the held object~\citep{wang2025unihope}.
The DINOv3 features capture the hand together with the held object, whereas the hand-pose backbone adds a strong prior on hand appearance.
The backbone features are resampled onto the $16\times16$ DINOv3 grid and concatenated channel-wise, and $\psi_h$ projects them with LayerNorm-linear layers~\citep{shi2025eagle}.
Each left hand is mirrored into a right hand by $\mathrm{diag}(-1,1,1)$ so that both hands share $\psi_h$ and the hand head, and its predictions are mirrored back.
A learned side embedding still distinguishes the two hands.
For the object, the frozen object-pose foundation model $E_o$~\citep{chen2026posegam} runs once per frame on $I$ and $V_o$.
The projection $\psi_o$ of Eqn.~\eqref{eq:inject} adds an MLP embedding of this feature to a learnable initial token $\mathbf{z}_o^{0}$.
The output layer of this MLP is zero-initialized, so training starts from $\mathbf{z}_o^{0}$ alone.
The transformer $\mathcal{F}$ comprises $L=12$ global self-attention blocks of width 768 and four register tokens~\citep{darcet2024registers}.

The hand and object heads are two-layer MLPs with hidden widths of 768 and 256, respectively.
The hand head predicts the MANO~\citep{romero2017mano} pose $\theta_h$ as 6D joint rotations~\citep{zhou2019continuity} and the shape $\beta_h$.
For the translation $\mathbf{t}_h$, it predicts a weak-perspective camera $(s_h, x_h, y_h)$~\citep{pavlakos2024hamer}, which the intrinsics $K$ lift to
\begin{equation}
\mathbf{t}_h = \Big(x_h,\ y_h,\ \frac{2f}{s_h W}\Big)^{\!\top},
\label{eq:cam}
\end{equation}
where $s_h$ is the weak-perspective scale, and $W$ and $f$ are the crop size and the focal length of $K$ in crop pixels.
The hand depth thus comes from the camera intrinsics rather than from the crop alone.
The object head predicts $\mathbf{R}_o$ in the same 6D representation and the translation offset $\Delta\mathbf{t}_o$ from the grasping hand $a$, so $\mathbf{t}_o=\mathbf{t}_a+\Delta\mathbf{t}_o$ needs no separate lift.

\vspace{0.5em}\noindent\textbf{Contact heads.}\quad
The coarse contact head of Eqn.~\eqref{eq:contact} predicts a contact probability for each of the $J=21$ hand joints from the hand and object tokens.
Its pooled feature $\hat{\mathbf{Z}}_h^{\mathrm{pool}}$ is obtained by cross-attention over $\hat{\mathbf{Z}}_h$ with a separate learned query for each hand, whereas the attention and the two-layer MLP $\phi_{\mathrm{c}}$ are shared across hands.
The predicted contact in turn informs the pose estimation, as two zero-initialized linear layers add it to $\hat{\mathbf{z}}_h$ and $\hat{\mathbf{z}}_o$ before the hand and object heads.
During training, this path also takes the predicted rather than the GT contact to match the input at inference.
Its gradient is stopped so that the pose losses never reach the contact head through this path.

The fine head operates on the finer hand regions~\citep{jung2026contactprompt}, each comprising 2 to 20 vertices.
They partition the palmar, dorsal, radial, and ulnar sides of the hand (\eg, finger segments, fingertips, the thenar, and the webspaces).
For region $p$ with vertices $\mathcal{V}_p$, the region distance $\mathbf{u}_{h,p}$ collects four statistics of the distances $d_{h,i}$ of Eqn.~\eqref{eq:regiondist},
\begin{equation}
\mathbf{u}_{h,p}=\left[\min_{i\in\mathcal{V}_p} d_{h,i},\ \frac{1}{|\mathcal{V}_p|}\sum_{i\in\mathcal{V}_p} d_{h,i},\ \frac{1}{|\mathcal{V}_p|}\sum_{i\in\mathcal{V}_p}\mathbbm{1}[d_{h,i}<\tau_{\mathrm{c}}],\ \frac{1}{|\mathcal{V}_p|}\sum_{i\in\mathcal{V}_p}\cos\big(\nabla d_{h,i},\mathbf{n}_{h,i}\big)\right],
\label{eq:finegeo}
\end{equation}
where $\tau_{\mathrm{c}}$ is the contact distance of the joint labels in Eqn.~\eqref{eq:contactloss}.
The first three entries measure how close the region is to the object and how much of it lies within this distance.
The last entry averages the cosine between the distance gradient $\nabla d_{h,i}$ and the outward hand normal $\mathbf{n}_{h,i}$, which is negative when the region faces the object and thus tells the two sides of a finger apart.
In $\mathcal{F}_{\mathrm{f}}$, each region token sums a learnable region embedding with embeddings of $\hat{\mathbf{c}}_h$ and $\mathbf{u}_{h,p}$, and one transformer layer of width 256 mixes the tokens.
A linear readout then predicts the contact logit and the signed distance $\hat{s}_{h,p}$ of each region.

\vspace{0.5em}\noindent\textbf{Training.}\quad
The first stage minimizes Eqn.~\eqref{eq:loss} and gradually adds the physical terms of Eqn.~\eqref{eq:stage2} after a warm-up.
Early stopping on the validation split then selects the best epoch among those with the physical terms active.
The second stage trains the fine head on the frozen model with the two equally weighted terms of Eqn.~\eqref{eq:fineloss}, again selecting the epoch on the validation split.
Since the first-stage model reconstructs its training images more accurately than unseen ones, the fine head would over-trust the region distances.
At every epoch, we thus shift the object in each training reconstruction by the offset error of Eqn.~\eqref{eq:rel} on a random validation image and recompute the region distances so that they carry test-like errors.
Both stages use AdamW~\citep{loshchilov2019adamw} with a cosine-decayed learning rate of $10^{-4}$, a weight decay of $0.05$, and a batch size of 256.
All results use an exponential moving average of the first-stage weights with a decay of $0.999$.
When one- and two-hand samples share a batch, the outputs and losses of a missing hand are discarded.

\subsection{Loss Terms}
\label{app:lossforms}
We provide the full formulations of the loss terms in Section~\ref{subsec:loss}, where $\rho$ is the smooth-$\ell_1$ loss and bars denote GT.

\vspace{0.5em}\noindent\textbf{Pose terms.}\quad
Each hand term $\mathcal{L}_h$ follows the standard hand supervision~\citep{pavlakos2024hamer} with an added translation term,
\begin{align}
\mathcal{L}_h &= \rho\big(\mathbf{t}_h - \bar{\mathbf{t}}_h\big)
+ \big\lVert \mathbf{J}_h - \bar{\mathbf{J}}_h \big\rVert_1\\
&\quad + \tfrac{1}{W}\big\lVert \pi_K(\mathbf{J}_h + \mathbf{t}_h) - \pi_K(\bar{\mathbf{J}}_h + \bar{\mathbf{t}}_h) \big\rVert_1\\
&\quad + \rho\big(R(\theta_h) - R(\bar{\theta}_h)\big) + \rho\big(\beta_h - \bar{\beta}_h\big)\\
&\quad + \lambda_{\mathrm{adv}}\big(D(\theta_h,\beta_h) - 1\big)^2,
\label{eq:handloss}
\end{align}
where $\mathbf{J}_h$ are the 21 wrist-relative 3D joints and $\bar{\mathbf{t}}_h$ is the GT wrist position.
The projection $\pi_K$ uses the crop intrinsics and $R(\theta)$ stacks the rotation matrices of the MANO pose.
A discriminator $D$ on the pose and shape acts as an adversarial prior with weight $\lambda_{\mathrm{adv}}=10^{-3}$.
For the object, $\mathcal{L}_o$ is the point-matching loss~\citep{xiang2018posecnn, wang2021gdrnet, mo2022es6d} on $M=64$ farthest-point samples $\mathbf{x}_k$ of $V_o$.
Its translation is taken relative to the grasping hand $a$ as in Eqn.~\eqref{eq:rel},
\begin{equation}
\mathcal{L}_o = \frac{1}{M}\sum_{k=1}^{M} \rho\Big(\big(\mathbf{R}_o - \bar{\mathbf{R}}_o\big)\mathbf{x}_k + \Delta\mathbf{t}_o - \big(\bar{\mathbf{t}}_o - \bar{\mathbf{t}}_a\big)\Big),
\label{eq:objloss}
\end{equation}
so that a hand depth error is not charged to the object.
The contact term of Eqn.~\eqref{eq:loss} is weighted by $\lambda_{\mathrm{c}}=0.1$.

\vspace{0.5em}\noindent\textbf{Object symmetry.}\quad
Each object term uses the symmetry-equivalent GT closest to the prediction in the symmetry group $\mathcal{G}_o$,
\begin{align}
\mathbf{G}^{\star} &= \argmin_{\mathbf{G}\in\mathcal{G}_o}\ d_{\mathrm{geo}}\big(\mathbf{R}_o,\ \bar{\mathbf{R}}_o\mathbf{G}\big),\\
\big(\bar{\mathbf{R}}_o,\ \bar{\mathbf{t}}_o\big) &\leftarrow \big(\bar{\mathbf{R}}_o\mathbf{G}^{\star},\ \bar{\mathbf{t}}_o + \bar{\mathbf{R}}_o(\boldsymbol{\mu}_o - \mathbf{G}^{\star}\boldsymbol{\mu}_o)\big),
\label{eq:sym}
\end{align}
where $d_{\mathrm{geo}}$ is the geodesic distance between rotations~\citep{huynh2009metrics} and $\boldsymbol{\mu}_o$ is the centroid of the mesh surface, about which the symmetries act.
The selected target is held constant, so the gradient pulls the prediction toward its nearest equivalent pose.
We detect each group by testing discrete rotations about the principal axes of the mesh, adding a continuous symmetry for near-cylindrical objects.
A candidate joins $\mathcal{G}_o$ when it maps the mesh onto itself within a Chamfer tolerance.
This criterion labels 18 of the 21 DexYCB objects and 78 of the 101 OakInk objects as symmetric.\nocite{hodan2024bop}

\vspace{0.5em}\noindent\textbf{Physical terms.}\quad
Each mesh $V_o$ has a signed distance function $\mathrm{SDF}_o$~\citep{park2019deepsdf}, precomputed on a $128^3$ grid in its frame.
We write $\Phi_o(\mathbf{x})=\mathrm{SDF}_o\big(\mathbf{R}_o^{\top}(\mathbf{x}-\mathbf{t}_o)\big)$ for this field under the predicted pose as in Eqn.~\eqref{eq:regiondist} and $\bar{\Phi}_o$ for that under the GT pose.
We use the object field because the MANO mesh is open at the wrist.
The first stage adds penetration and attraction terms to Eqn.~\eqref{eq:loss},
\begin{equation}
\mathcal{L}_{\mathrm{phys}} = \mathcal{L}
+ \sum_{h}\big(\lambda_{\mathrm{pen}}\,\mathcal{L}^{h}_{\mathrm{pen}}
+ \lambda_{\mathrm{att}}\,\mathcal{L}^{h}_{\mathrm{att}}\big),
\label{eq:stage2}
\end{equation}
with $\lambda_{\mathrm{pen}}=100$ and $\lambda_{\mathrm{att}}=10$.
The penetration term penalizes interpenetration in both directions,
\begin{equation}
\mathcal{L}^{h}_{\mathrm{pen}} = \frac{1}{|\mathcal{V}_h|}\sum_{i\in\mathcal{V}_h}\max\big(0,\,-\Phi_o(\mathbf{v}_{h,i})\big)
+ \frac{1}{|\mathcal{S}|}\sum_{\mathbf{p}\in\mathcal{S}}\mathbbm{1}\big[\delta_h(\mathbf{p})<0\big]\,\big\lVert\mathbf{p}-\mathbf{v}_h(\mathbf{p})\big\rVert_2,
\label{eq:pen}
\end{equation}
where $\mathcal{V}_h$ indexes the predicted vertices $\mathbf{v}_{h,i}$ of hand $h$, $\mathcal{S}$ is a set of points sampled on the predicted object surface, and $\mathbbm{1}[\cdot]$ is the indicator function.
The first sum penalizes each hand vertex by its depth inside the object.
The second penalizes each object point by its distance to the nearest hand vertex $\mathbf{v}_h(\mathbf{p})$ when the offset $\delta_h(\mathbf{p})=(\mathbf{p}-\mathbf{v}_h(\mathbf{p}))^{\top}\mathbf{n}_h(\mathbf{p})$ along the outward normal of that vertex is negative (\ie, $\mathbf{p}$ lies inside the hand).
The attraction term~\citep{hasson2019obman, grady2021contactopt} acts on the GT contact set $\mathcal{C}_h$ of hand vertices within $3\,\mathrm{mm}$ of the object, following the CDev rule~\citep{fan2023arctic}.
It matches the signed distance of each such vertex to its GT value and weights it by the predicted contact of its joint $j(i)$,
\begin{equation}
\mathcal{L}^{h}_{\mathrm{att}} = \frac{\sum_{i\in\mathcal{C}_h} w_{h,i}\,\big|\Phi_o(\mathbf{v}_{h,i}) - \bar{\Phi}_o(\bar{\mathbf{v}}_{h,i})\big|}{\sum_{i\in\mathcal{C}_h} w_{h,i}},
\label{eq:att}
\end{equation}
where a stop-gradient on $w_{h,i}=\hat{c}_{h,j(i)}$ keeps the term from lowering the predicted contact.

\vspace{0.5em}\noindent\textbf{Contact term.}\quad
The joint contact labels follow the signed-distance rule of HACO~\citep{jung2025haco} under the GT poses.
A joint is in contact when at least 5\% of the vertices $\mathcal{V}_j$ assigned to it in the rest pose~\citep{hasson2019obman} lie within $\tau_{\mathrm{c}}$ of the object or inside it,
\begin{equation}
c^{\star}_{h,j} = \mathbbm{1}\Big[\,\big|\{\,i\in\mathcal{V}_j : \bar{\Phi}_o(\bar{\mathbf{v}}_{h,i}) < \tau_{\mathrm{c}}\,\}\big| \geq 0.05\,|\mathcal{V}_j|\,\Big].
\label{eq:contactloss}
\end{equation}
Following HACO, we set $\tau_{\mathrm{c}}=1\,\mathrm{cm}$ because annotated hands can sit up to $5\,\mathrm{mm}$ inside their objects and a tighter threshold would turn this offset into label noise.
The coarse contact loss $\mathcal{L}^{h}_{\mathrm{c}}$ is the class-balanced objective of HACO,
\begin{align}
\mathcal{L}^{h}_{\mathrm{c}} &= \frac{1}{J}\sum_{j} w_{j,c^{\star}_{h,j}}\,\mathrm{BCE}\big(\hat{c}_{h,j},\,c^{\star}_{h,j}\big)
+ \frac{\lambda_{\mathrm{fp}}}{|\mathcal{N}_h|}\sum_{j\in\mathcal{N}_h}\hat{c}_{h,j}\\
&\quad + \frac{\lambda_{\mathrm{pr}}}{J}\sum_{j}\big|\hat{c}_{h,j} - c^{\mathrm{pr}}_j\big|\\
&\quad + \lambda_{\mathrm{sm}}\,\log\Big(1 + \frac{2}{J}\sum_{j}\big|\hat{c}_{h,j} - \langle\hat{\mathbf{c}}_h\rangle_{\mathcal{A}(j)}\big|\Big).
\label{eq:vcb}
\end{align}
The first term is a binary cross-entropy weighted by $w_{j,y} \propto (1-\gamma)/(1-\gamma^{n_{j,y}})$ with $\gamma=0.9999$~\citep{cui2019class}, where $n_{j,y}$ counts label $y$ at joint $j$ per 1,000 samples.
As in HACO, $w_{j,y}$ is normalized per joint and annealed from a global toward a per-joint class balance.
The second term penalizes predicted contact on the non-contact joints $\mathcal{N}_h$.
The third term pulls each prediction toward the training contact rate $c^{\mathrm{pr}}_j$ of its joint.
The last term discourages isolated decisions by comparing each joint with the mean $\langle\hat{\mathbf{c}}_h\rangle_{\mathcal{A}(j)}$ of its skeleton neighbors $\mathcal{A}(j)$.
The weights $(\lambda_{\mathrm{fp}},\lambda_{\mathrm{pr}},\lambda_{\mathrm{sm}}) = (0.5, 0.1, 1)$ follow HACO.

\subsection{Datasets}
\label{app:data}
For one-hand interactions, we use the official s0 and s3 splits of DexYCB~\citep{chao2021dexycb}, the official HO3Dv2 partition~\citep{hampali2020honnotate}, and the rigid-object categories of HOI4D~\citep{liu2022hoi4d}.
We also use the official split0 of OakInk~\citep{yang2022oakink}, which holds out one camera view of each sequence.
For two-hand interactions, we use the official H2O split~\citep{kwon2021h2o} and a sequence-level split of OakInk2~\citep{zhan2024oakink2}.
A closure filter keeps the one-hand training frames in which at least four hand vertices lie within 5\,mm of the object and their contacts are in force closure~\citep{liu2021synthesizing}.
This removes frames in which the hand approaches, releases, or unstably holds the object.
Two visibility filters then require at least four of the 21 joints and 90\% of the object vertices to project into the image.
\mbox{\methodall} is trained once with the same architecture and recipe on the union of all training splits, excluding every test frame.
One- and two-hand samples from different datasets thus jointly train this single model.

\subsection{Evaluation Protocol and Metrics}
\label{subsec:protocol}
Following \mbox{UniHOPE}~\citep{wang2025unihope}, the one-hand test sets keep the grasping frames where the object has left its initial pose.
All methods use GT localization so that detection errors do not affect the comparison.

\vspace{0.5em}\noindent\textbf{Hand pose.}\quad
MJE and PA-MJE average the errors of the 21 joints $\{\mathbf{J}_j\}_{j\in\mathcal{J}}$ (wrist $\mathbf{J}_0$) after wrist and Procrustes alignment,
\begin{align}
\mathrm{MJE} &= \frac{1}{|\mathcal{J}|}\sum_{j\in\mathcal{J}}\big\lVert(\mathbf{J}_j-\mathbf{J}_0)-(\bar{\mathbf{J}}_j-\bar{\mathbf{J}}_0)\big\rVert_2,\\
\mathrm{PA\text{-}MJE} &= \frac{1}{|\mathcal{J}|}\sum_{j\in\mathcal{J}}\big\lVert s^{\star}\mathbf{R}^{\star}\mathbf{J}_j+\mathbf{t}^{\star}-\bar{\mathbf{J}}_j\big\rVert_2,
\label{eq:mje}
\end{align}
where $(s^{\star},\mathbf{R}^{\star},\mathbf{t}^{\star})$ is the similarity transform minimizing $\sum_{j}\lVert s\mathbf{R}\mathbf{J}_j+\mathbf{t}-\bar{\mathbf{J}}_j\rVert_2^2$.
MJE thus reflects the finger articulation along with the global orientation and size of the hand, whereas PA-MJE isolates the articulation.
VPE and PA-VPE compute the same errors over the 778 MANO vertices.
For two hands, MRRPE~\citep{fan2023arctic} measures the error $\lVert(\mathbf{J}^{l}_0-\mathbf{J}^{r}_0)-(\bar{\mathbf{J}}^{l}_0-\bar{\mathbf{J}}^{r}_0)\rVert_2$ of the left-to-right wrist offset.
Predicted and GT joints are regressed from the MANO vertices by the same regressor.

\vspace{0.5em}\noindent\textbf{Object pose.}\quad
ADD~\citep{hinterstoisser2012model} averages the camera-frame distance between the points $\mathcal{X}$ of $V_o$ under the predicted and GT poses,
\begin{equation}
\mathrm{ADD} = \frac{1}{|\mathcal{X}|}\sum_{\mathbf{x}\in\mathcal{X}}\big\lVert(\mathbf{R}_o\mathbf{x}+\mathbf{t}_o)-(\bar{\mathbf{R}}_o\mathbf{x}+\bar{\mathbf{t}}_o)\big\rVert_2.
\label{eq:add}
\end{equation}
Following \mbox{UniHOPE}~\citep{wang2025unihope}, the one-hand test sets score symmetric objects and objects without a symmetry annotation with \mbox{ADD-S}~\citep{xiang2018posecnn}.
\mbox{ADD-S} matches each predicted point to its closest GT point and thus accepts symmetric poses.
The AUC of ADD-0.5D is the percentage of the $N$ test frames whose error $e_n$ lies within half the object diameter $D_o$,
\begin{equation}
\mathrm{AUC} = \frac{100}{N}\sum_{n=1}^{N}\mathbbm{1}\big[e_n\le 0.5\,D_o\big].
\label{eq:auc}
\end{equation}
A higher AUC thus indicates that more frames have an accurate object pose.
Since \method\ places the object relative to the predicted hand, its object errors also include the hand translation error.

\vspace{0.5em}\noindent\textbf{Joint pose.}\quad
SD and IV~\citep{hasson2019obman} measure the physical plausibility of the predicted hand and object.
SD fixes the hand, simulates the object under gravity in PyBullet~\citep{coumans2016pybullet}, and reports its center displacement in centimeters.
IV reports the volume of the 1\,cm object voxels inside the hand mesh in cubic centimeters.
As IV also drops when the object moves away, we add CDev~\citep{fan2023arctic, bansal2026hopformer}.
It averages the distance to the object from each predicted hand vertex in the GT contact set $\mathcal{C}_h$ of Eqn.~\eqref{eq:att},
\begin{equation}
\mathrm{CDev} = \frac{1}{|\mathcal{C}_h|}\sum_{i\in\mathcal{C}_h}\min_{\mathbf{y}\in\mathcal{V}_o}\big\lVert\mathbf{v}_{h,i}-\mathbf{y}\big\rVert_2,
\label{eq:cdev}
\end{equation}
where $\mathcal{V}_o$ are the vertices of $V_o$ under the predicted pose, and the nearest vertex makes CDev robust to object symmetry.
A PnP~\citep{lepetit2009epnp} or RANSAC~\citep{fischler1981random} failure can place a baseline's object meters away, so CDev omits frames where it exceeds 1\,m.

\vspace{0.5em}\noindent\textbf{Contact.}\quad
A joint is predicted in contact when $\hat{c}_{h,j}>0.5$, and the F1 score is pooled over all joints and frames,
\begin{equation}
\mathrm{F1} = \frac{2\,\mathrm{TP}}{2\,\mathrm{TP}+\mathrm{FP}+\mathrm{FN}}.
\label{eq:f1}
\end{equation}
F1 thus penalizes both spurious and missed contact.
F1@1cm uses the labeling rule of HACO~\citep{jung2025haco} in Eqn.~\eqref{eq:contactloss}.
F1@3mm uses the stricter 3\,mm rule of CDev~\citep{fan2023arctic}, which replaces $\bar{\Phi}_o<\tau_{\mathrm{c}}$ with $|\bar{\Phi}_o|<3\,\mathrm{mm}$.
The fine contact $\tilde{c}_{h,p}$ is scored likewise over the finer regions.
The joint estimators are scored by these rules on their predicted meshes.
For HACO, the probability of joint $j$ is the $\lceil 0.05\,|\mathcal{V}_j| \rceil$-th largest vertex probability in $\mathcal{V}_j$ so that a 0.5 threshold applies Eqn.~\eqref{eq:contactloss} to its vertex decisions.
The expected calibration error (ECE) is computed over $B=10$ equal-width bins $\mathcal{B}_b$ of the $N_{\mathrm{c}}$ joint probabilities,
\begin{equation}
\mathrm{ECE} = \sum_{b=1}^{B}\frac{|\mathcal{B}_b|}{N_{\mathrm{c}}}\,\big|\mathrm{acc}(\mathcal{B}_b)-\mathrm{conf}(\mathcal{B}_b)\big|,
\label{eq:ece}
\end{equation}
where $\mathrm{acc}$ and $\mathrm{conf}$ are the empirical contact rate and the mean probability of a bin, which coincide for a calibrated model.

\section{Additional Results}
\label{app:analyses}
This section provides additional quantitative results of \method\ that extend the evaluations in Section~\ref{sec:evaluations}.
We first compare \method\ with the baselines on unseen objects and under further metrics in Supp.~\ref{app:full_transfer}.
We then analyze the robustness to occlusion in Supp.~\ref{app:full_tables}.
Finally, we ablate the frozen foundation features of the hand and object in Supp.~\ref{app:ablation}.

\subsection{Complete Comparison Tables}
\label{app:full_transfer}
\textbf{In-domain unseen objects.}\quad
\label{subsec:base2novel}
The unseen objects of OakInk and HOI4D in Table~\ref{tab:domaingen} also come with new cameras and scenes.
We therefore isolate object novelty on the official DexYCB s3 split, whose three test objects are absent from training and given only as meshes.
As shown in Table~\ref{tab:basenovel}, \method\ nearly doubles the object AUC of the strongest joint estimator from 43.1 to 85.2.
It also attains the lowest MJE and PA-MJE of 10.5 and 5.1 against at least 13.9 and 6.1 for the baselines.
MegaPose reaches a higher average AUC of 87.3 owing to the small gelatin box, while \method\ leads on the other two objects with 97.9 and 99.8.
Yet MegaPose estimates the object apart from the hand and leaves a CDev of about 41 with any hand estimator against 15.5 for \method.
The joint estimators leave the object even farther away with a CDev above 120, which mainly accounts for their lower IV.
These results indicate that object novelty alone limits the joint estimators, whereas encoding the object from its geometry carries over to unseen objects.
\begin{table}[!h]
\centering
\caption{In-domain comparison on the official DexYCB s3 split, whose three test objects are unseen in training and given only as meshes. \cmb~denotes pairing with the MegaPose object (highest AUC) or the H2ONet hand (lowest PA-MJE). \method\ nearly doubles the object AUC of the strongest joint estimator and attains the lowest hand errors and CDev of all methods.}
\label{tab:basenovel}
\par\vspace{-0.5em}
\resizebox{\linewidth}{!}{%
\begin{tabular}{l|cccc|cccc|ccc}
\toprule
& \multicolumn{4}{c|}{\textbf{Object Pose}} & \multicolumn{4}{c|}{\textbf{Hand Pose}} & \multicolumn{3}{c}{\textbf{Joint Pose}} \\
\cmidrule(lr){2-5}\cmidrule(lr){6-9}\cmidrule(lr){10-12}
\textbf{Method} & \multicolumn{4}{c|}{AUC of ADD-0.5D ($\uparrow$)} & MJE ($\downarrow$) & PA-MJE ($\downarrow$) & VPE ($\downarrow$) & PA-VPE ($\downarrow$) & SD ($\downarrow$) & IV ($\downarrow$) & CDev ($\downarrow$) \\
\midrule
\textbf{\underline{Hand Pose Estimation}} & \textcolor{gray}{\textit{gelatin\_box}} & \textcolor{gray}{\textit{bleach\_cleanser}} & \textcolor{gray}{\textit{wood\_block}} & \textcolor{gray}{\textit{average}} & & & & & & & \\
HandOccNet~\citep{park2022handoccnet} & \na & \na & \na & \na & 14.58 & 6.73 & 14.10 & 6.49 & 4.49\cmb & 16.63\cmb & \phantom{0}41.24\cmb \\
MobRecon~\citep{chen2022mobrecon} & \na & \na & \na & \na & 15.40 & 7.25 & 14.24 & 6.39 & \underline{4.48}\cmb & 17.09\cmb & \phantom{0}\underline{40.93}\cmb \\
H2ONet~\citep{xu2023h2onet} & \na & \na & \na & \na & 15.20 & 6.35 & 15.03 & 6.74 & 4.67\cmb & 13.38\cmb & \phantom{0}41.24\cmb \\
SimpleHand~\citep{zhou2024simplehand} & \na & \na & \na & \na & 14.88 & 6.74 & 14.21 & 6.45 & \textbf{4.42}\cmb & 15.68\cmb & \phantom{0}40.99\cmb \\
\midrule
\textbf{\underline{Object Pose Estimation}} & \textcolor{gray}{\textit{gelatin\_box}} & \textcolor{gray}{\textit{bleach\_cleanser}} & \textcolor{gray}{\textit{wood\_block}} & \textcolor{gray}{\textit{average}} & & & & & & & \\
MegaPose~\citep{labbe2022megapose} & \textbf{82.97} & \underline{88.61} & 90.80 & \textbf{87.27} & \na & \na & \na & \na & 4.67\cmb & 13.38\cmb & \phantom{0}41.24\cmb \\
GigaPose~\citep{nguyen2024gigapose} & 60.89 & 73.48 & \underline{92.81} & 74.94 & \na & \na & \na & \na & 6.83\cmb & 12.22\cmb & \phantom{0}56.47\cmb \\
FoundPose~\citep{ornek2024foundpose} & 54.42 & 84.23 & 85.94 & 74.14 & \na & \na & \na & \na & 6.70\cmb & 12.52\cmb & \phantom{0}59.62\cmb \\
PoseGAM~\citep{chen2026posegam} & \underline{69.59} & 73.14 & 82.10 & 74.63 & \na & \na & \na & \na & 7.04\cmb & \phantom{0}\textbf{8.62}\cmb & \phantom{0}79.96\cmb \\
\midrule
\textbf{\underline{Hand Object Pose Estimation}} & \textcolor{gray}{\textit{gelatin\_box}} & \textcolor{gray}{\textit{bleach\_cleanser}} & \textcolor{gray}{\textit{wood\_block}} & \textcolor{gray}{\textit{average}} & & & & & & & \\
SemiHand~\citep{liu2021semihand} & 25.96 & 25.45 & 68.69 & 38.93 & 15.35 & 6.51 & 14.92 & 6.54 & 8.37 & \phantom{0}\underline{9.71} & 131.95 \\
HFL-Net~\citep{lin2023hfl} & 27.23 & 29.73 & 69.75 & 41.15 & 15.31 & 6.63 & 14.81 & 6.61 & 8.12 & 10.65 & 122.00 \\
UniHOPE~\citep{wang2025unihope} & 26.41 & 32.64 & 73.88 & 43.11 & \underline{13.92} & \underline{6.07} & \underline{13.49} & \underline{6.05} & 8.09 & 10.93 & 121.72 \\
\textbf{\method\ (ours)} & 60.49 & \textbf{97.91} & \textbf{99.83} & \underline{85.17} & \textbf{10.51} & \textbf{5.05} & \textbf{10.15} & \textbf{5.04} & 4.76 & 15.25 & \phantom{0}\textbf{15.53} \\
\bottomrule
\end{tabular}}
\end{table}

\vspace{1em}
\begin{wraptable}{r}{0.45\linewidth}
\vspace{-1.2em}
\centering
\caption{Wrist-aligned hand errors in mm for the cross-dataset transfer of Table~\ref{tab:domaingen}. \method\ halves the MJE of the joint estimators on HO3D and HOI4D, far beyond its PA-MJE gain.}
\label{tab:domaingen_full}
\par\vspace{-0.5em}
\resizebox{\linewidth}{!}{%
\begin{tabular}{l|cc|cc|cc}
\toprule
& \multicolumn{2}{c|}{\textbf{OakInk}~\citep{yang2022oakink}} & \multicolumn{2}{c|}{\textbf{HO3D}~\citep{hampali2020honnotate}} & \multicolumn{2}{c}{\textbf{HOI4D}~\citep{liu2022hoi4d}} \\
\cmidrule(lr){2-3}\cmidrule(lr){4-5}\cmidrule(lr){6-7}
\textbf{Method} & MJE ($\downarrow$) & VPE ($\downarrow$) & MJE ($\downarrow$) & VPE ($\downarrow$) & MJE ($\downarrow$) & VPE ($\downarrow$) \\
\midrule
SemiHand~\citep{liu2021semihand} & 22.47 & 21.83 & 44.84 & 43.10 & 49.34 & 47.90 \\
HFL-Net~\citep{lin2023hfl} & 25.16 & 24.42 & 43.31 & 41.76 & 64.90 & 62.90 \\
UniHOPE~\citep{wang2025unihope} & 27.28 & 26.45 & 43.95 & 42.36 & 49.16 & 47.93 \\
\textbf{\method\ (ours)} & \underline{16.02} & \underline{15.62} & \underline{22.22} & \underline{21.54} & \underline{23.64} & \underline{23.30} \\
\textbf{\methodall\ (ours)} & \phantom{0}\textbf{9.19} & \phantom{0}\textbf{8.96} & \textbf{20.07} & \textbf{19.46} & \textbf{12.63} & \textbf{12.28} \\
\bottomrule
\end{tabular}}
\vspace{-1.5em}
\end{wraptable}
\noindent\textbf{Additional hand metrics.}\quad
Table~\ref{tab:domaingen} reports Procrustes-aligned hand errors, which isolate the finger articulation.
Table~\ref{tab:domaingen_full} adds the wrist-aligned errors, which also reflect the global orientation and size of the hand.
\mbox{\method} roughly halves the MJE of the joint estimators on HO3D and HOI4D and lowers it by at least 6.4 on OakInk.
\mbox{\methodall} lowers it further on every target.
In contrast, the PA-MJE gain of \method\ in Table~\ref{tab:domaingen} is at most 1.4.
Most of the transfer gain thus lies in the global orientation and size, which Procrustes alignment hides.

\begin{table}[t]
\centering
\caption{Region-level contact F1 over all finer hand regions~\citep{jung2026contactprompt} and the 27 palm regions. HACO is pooled from its per-vertex decisions, and the other rows follow Table~\ref{tab:finecontact}. The fine contact of \method\ outperforms HACO in every setting, most clearly on the palm.}
\label{tab:finecontact_full}
\par\vspace{-0.5em}
\resizebox{\linewidth}{!}{%
\begin{tabular}{l|cc|cc|cc|cc}
\toprule
& \multicolumn{4}{c|}{\textbf{DexYCB}~\citep{chao2021dexycb}} & \multicolumn{4}{c}{\textbf{OakInk}~\citep{yang2022oakink}} \\
\cmidrule(lr){2-5}\cmidrule(lr){6-9}
& \multicolumn{2}{c}{All} & \multicolumn{2}{c|}{Palm} & \multicolumn{2}{c}{All} & \multicolumn{2}{c}{Palm} \\
\cmidrule(lr){2-3}\cmidrule(lr){4-5}\cmidrule(lr){6-7}\cmidrule(lr){8-9}
\textbf{Method} & F1@1cm ($\uparrow$) & F1@3mm ($\uparrow$) & F1@1cm ($\uparrow$) & F1@3mm ($\uparrow$) & F1@1cm ($\uparrow$) & F1@3mm ($\uparrow$) & F1@1cm ($\uparrow$) & F1@3mm ($\uparrow$) \\
\midrule
HACO~\citep{jung2025haco} & 77.87 & 52.95 & 40.91 & 19.14 & 68.88 & 42.80 & 33.59 & 13.90 \\
\midrule
Coarse only & 78.97 & 54.92 & 46.28 & 25.91 & 77.49 & 49.21 & 48.42 & 29.92 \\
Fine only & \textbf{80.75} & \underline{60.98} & 46.30 & 37.52 & 80.65 & 56.13 & 62.28 & 44.26 \\
Coarse-to-fine & 80.37 & 60.44 & \textbf{50.21} & \underline{37.64} & \underline{81.55} & \underline{56.48} & \textbf{64.93} & \underline{47.14} \\
\textit{+ geometry (ours)} & \underline{80.59} & \textbf{61.63} & \underline{49.82} & \textbf{39.97} & \textbf{82.32} & \textbf{57.58} & \underline{64.22} & \textbf{50.49} \\
\bottomrule
\end{tabular}}
\end{table}
\vspace{1em}\noindent\textbf{Region-level contact.}\quad
The main paper compares the fine contact only with its own variants.
We therefore add HACO~\citep{jung2025haco}, the strongest contact baseline.
Since HACO predicts contact per vertex, a region counts as HACO contact when any of its vertices exceeds the decision threshold.
As shown in Table~\ref{tab:finecontact_full}, the fine contact of \method\ outperforms this pooled HACO under both rules on both datasets.
The margin is largest on the palm regions, where the F1@3mm rises from 19.1 to 40.0 on DexYCB and from 13.9 to 50.5 on OakInk.
Among the variants, our model ranks first or second in every setting, and the region distances raise the palm F1@3mm of \emph{Coarse-to-fine} by 2.3 and 3.4 points.
These results indicate that the reconstruction localizes contact more precisely than image features alone, most clearly on the palm.

\begingroup\clubpenalty=10000
\vspace{1em}\noindent\textbf{Stricter object metrics.}\quad
The AUC of ADD-0.5D~\citep{wang2025unihope} in the main tables accepts errors up to half the object diameter, and the closest-point ADD-S~\citep{xiang2018posecnn} also accepts a wrong rotation of a nearly symmetric object.
We therefore score every object with both the plain ADD and the ADD-S at 0.5, 0.3, and 0.1 of the diameter.
As shown in Table~\ref{tab:strictobj}, \method\ leads the joint estimators in five of the six average settings.
\mbox{\methodall} leads in all six.
The margin is largest on the unseen objects of OakInk and HOI4D, where the ADD-0.5D of \method\ is 56.4 and 26.6 against at most 11.9 and 3.0.
The only exception in the averages is the plain ADD at 0.1 of the diameter, where SemiHand leads on DexYCB and HO3D.
SemiHand recovers the object depth from the image by fitting the bounding box of the given mesh to its predicted keypoints with PnP~\citep{lepetit2009epnp}, whereas \method\ places the object relative to the predicted hand and inherits its depth error.
Aligning the predicted wrist to the GT wrist indeed raises this score of \method\ on HO3D from 4.0 to 23.1.
On DexYCB the remaining gap lies in rotation, as the ADD-S at 0.1 of the diameter reverses the order with 79.2 against 57.8.
These results indicate that the gains of \method\ hold under stricter metrics, whereas its finest object accuracy is bounded by the depth of the predicted hand.
\begin{table}[!h]
\centering
\caption{Object pose of the DexYCB-s0 models under stricter metrics, in domain and on the transfer targets. Each cell is the percentage of frames whose plain ADD (top) or closest-point ADD-S (bottom) lies within 0.5, 0.3, or 0.1 of the object diameter $D_o$. Average is the mean over the four datasets, and each block ends with the feature ablations. The DexYCB poses of the joint estimators come from a rerun of their RANSAC~\citep{fischler1981random} PnP~\citep{lepetit2009epnp} and differ from Table~\ref{tab:indomainpose} by at most 0.2 AUC. \method\ leads the joint estimators in five of the six average settings.}
\label{tab:strictobj}
\par\vspace{-0.5em}
\resizebox{\linewidth}{!}{%
\begin{tabular}{l|ccc|ccc|ccc|ccc|ccc}
\toprule
& \multicolumn{3}{c|}{\textbf{DexYCB}~\citep{chao2021dexycb}} & \multicolumn{3}{c|}{\textbf{HO3D}~\citep{hampali2020honnotate}} & \multicolumn{3}{c|}{\textbf{OakInk}~\citep{yang2022oakink}} & \multicolumn{3}{c|}{\textbf{HOI4D}~\citep{liu2022hoi4d}} & \multicolumn{3}{c}{\textbf{Average}} \\
\cmidrule(lr){2-4}\cmidrule(lr){5-7}\cmidrule(lr){8-10}\cmidrule(lr){11-13}\cmidrule(lr){14-16}
\textbf{Method} & 0.5\,$D_o$ & 0.3\,$D_o$ & 0.1\,$D_o$ & 0.5\,$D_o$ & 0.3\,$D_o$ & 0.1\,$D_o$ & 0.5\,$D_o$ & 0.3\,$D_o$ & 0.1\,$D_o$ & 0.5\,$D_o$ & 0.3\,$D_o$ & 0.1\,$D_o$ & 0.5\,$D_o$ & 0.3\,$D_o$ & 0.1\,$D_o$ \\
\midrule
\rowcolor{tabgray}\multicolumn{16}{c}{ADD} \\
\midrule
SemiHand~\citep{liu2021semihand} & 77.36 & \underline{59.42} & \textbf{25.46} & 86.81 & 67.81 & \textbf{28.10} & 11.93 & \phantom{0}2.50 & \phantom{0}0.03 & \phantom{0}2.70 & \phantom{0}0.34 & \phantom{0}0.00 & 44.70 & 32.52 & \underline{13.40} \\
HFL-Net~\citep{lin2023hfl} & 70.38 & 49.90 & \underline{16.03} & 81.16 & 56.97 & \underline{18.53} & 11.45 & \phantom{0}2.88 & \phantom{0}0.04 & \phantom{0}2.20 & \phantom{0}0.53 & \phantom{0}0.00 & 41.30 & 27.57 & \phantom{0}8.65 \\
UniHOPE~\citep{wang2025unihope} & 69.05 & 46.48 & 13.33 & 79.57 & 55.07 & 14.64 & 11.75 & \phantom{0}2.48 & \phantom{0}0.04 & \phantom{0}2.98 & \phantom{0}0.43 & \phantom{0}\underline{0.03} & 40.84 & 26.12 & \phantom{0}7.01 \\
\textbf{\method\ (ours)} & \underline{87.99} & 57.12 & 11.43 & \underline{88.75} & \underline{68.13} & \phantom{0}3.96 & \underline{56.40} & \underline{18.32} & \phantom{0}\underline{0.24} & \underline{26.58} & \phantom{0}\underline{5.24} & \phantom{0}\underline{0.03} & \underline{64.93} & \underline{37.20} & \phantom{0}3.92 \\
\textbf{\methodall\ (ours)} & \textbf{89.03} & \textbf{65.17} & 11.50 & \textbf{97.31} & \textbf{81.62} & 16.53 & \textbf{94.19} & \textbf{78.73} & \textbf{19.26} & \textbf{90.20} & \textbf{68.08} & \phantom{0}\textbf{7.54} & \textbf{92.68} & \textbf{73.40} & \textbf{13.71} \\
\midrule
\method\ w/o Object Feature & 71.84 & 27.15 & \phantom{0}2.10 & 78.41 & 44.16 & \phantom{0}0.61 & 45.97 & 10.28 & \phantom{0}0.09 & 19.20 & \phantom{0}4.59 & \phantom{0}0.00 & 53.85 & 21.54 & \phantom{0}0.70 \\
\method\ w/o Hand Feature & 84.16 & 48.81 & \phantom{0}8.15 & 86.02 & 65.15 & \phantom{0}4.45 & 51.12 & 17.96 & \phantom{0}0.24 & 22.43 & \phantom{0}3.69 & \phantom{0}0.00 & 60.93 & 33.90 & \phantom{0}3.21 \\
\midrule
\rowcolor{tabgray}\multicolumn{16}{c}{ADD-S} \\
\midrule
SemiHand~\citep{liu2021semihand} & 89.77 & 84.23 & 57.78 & \underline{97.32} & 94.25 & \textbf{69.51} & 40.50 & 27.83 & \phantom{0}3.42 & 15.57 & \phantom{0}9.93 & \phantom{0}1.33 & 60.79 & 54.06 & 33.01 \\
HFL-Net~\citep{lin2023hfl} & 85.56 & 78.21 & 48.50 & 94.72 & 91.64 & 66.23 & 37.83 & 26.32 & \phantom{0}3.49 & 13.52 & \phantom{0}8.22 & \phantom{0}0.84 & 57.91 & 51.10 & 29.77 \\
UniHOPE~\citep{wang2025unihope} & 85.27 & 77.49 & 44.91 & 94.22 & 90.26 & 60.91 & 38.81 & 27.00 & \phantom{0}3.17 & 14.17 & \phantom{0}8.68 & \phantom{0}1.40 & 58.12 & 50.86 & 27.60 \\
\textbf{\method\ (ours)} & \underline{99.08} & \underline{98.26} & \textbf{79.19} & \textbf{100.00} & \textbf{99.77} & 56.59 & \underline{93.29} & \underline{72.48} & \phantom{0}\underline{5.48} & \underline{67.68} & \underline{47.05} & \phantom{0}\underline{1.43} & \underline{90.01} & \underline{79.39} & \underline{35.67} \\
\textbf{\methodall\ (ours)} & \textbf{99.12} & \textbf{98.33} & \underline{76.33} & \textbf{100.00} & \underline{98.66} & \underline{68.61} & \textbf{99.97} & \textbf{99.11} & \textbf{54.54} & \textbf{97.36} & \textbf{93.55} & \textbf{29.22} & \textbf{99.11} & \textbf{97.41} & \textbf{57.17} \\
\midrule
\method\ w/o Object Feature & 99.05 & 98.06 & 69.54 & 99.95 & 99.12 & 46.78 & 94.08 & 73.64 & \phantom{0}3.25 & 59.06 & 34.15 & \phantom{0}0.81 & 88.04 & 76.24 & 30.09 \\
\method\ w/o Hand Feature & 98.66 & 95.97 & 67.93 & 99.87 & 96.29 & 53.10 & 89.09 & 68.47 & \phantom{0}5.79 & 63.21 & 39.95 & \phantom{0}0.43 & 87.71 & 75.17 & 31.82 \\
\bottomrule
\end{tabular}}
\end{table}
\vspace{-1.0em}
\FloatBarrier\par\endgroup

\subsection{Robustness to Occlusion}
\label{app:full_tables}
In a grasp, the hand and object hide each other's image evidence.
Following~\citet{wang2025unihope}, we bin the test frames by the fraction of the hand or object hidden by the other.
The DexYCB-s0 models are scored on DexYCB and on OakInk without retraining.
Table~\ref{tab:handocc} shows that every method degrades toward the most occluded hand bin on DexYCB.
\method\ nevertheless leads every baseline in every bin, most clearly in the middle bin with a PA-MJE of 5.18 against 6.00.
On OakInk the hand errors show no occlusion trend and \method\ stays below every baseline.
Occlusion harms the object far more.
On DexYCB the AUC of the joint estimators falls from between 78.70 and 86.96 in the first bin to between 18.22 and 25.44 in the last, whereas \method\ keeps 93.51 and 72.65.
On OakInk \method\ stays above 90, whereas the joint estimators reach only 23.49 to 41.35 as they fail on the unseen objects.
Only \method\ thus remains accurate under heavy object occlusion.
\begin{table}[!h]
\centering
\caption{Robustness to occlusion of the DexYCB-s0 models on DexYCB and on OakInk without retraining. Frames are binned by the fraction of the hand or object hidden by the other. We report PA-MJE and PA-VPE in mm under hand occlusion and the object AUC under object occlusion. \method\ leads the baselines in every bin, most clearly under heavy object occlusion.}
\label{tab:handocc}
\label{tab:objocc}
\par\vspace{-0.5em}
\resizebox{\linewidth}{!}{%
\begin{tabular}{l|cccccc|ccc|cccccc|ccc}
\toprule
& \multicolumn{9}{c|}{\textbf{Source Dataset: DexYCB}~\citep{chao2021dexycb}} & \multicolumn{9}{c}{\textbf{Target Dataset: OakInk}~\citep{yang2022oakink}} \\
\cmidrule(lr){2-10}\cmidrule(lr){11-19}
& \multicolumn{6}{c|}{Hand Occlusion (PA-MJE $\downarrow$ / PA-VPE $\downarrow$)} & \multicolumn{3}{c|}{Object Occlusion (AUC $\uparrow$)} & \multicolumn{6}{c|}{Hand Occlusion (PA-MJE $\downarrow$ / PA-VPE $\downarrow$)} & \multicolumn{3}{c}{Object Occlusion (AUC $\uparrow$)} \\
\cmidrule(lr){2-7}\cmidrule(lr){8-10}\cmidrule(lr){11-16}\cmidrule(lr){17-19}
\textbf{Method} & \multicolumn{2}{c}{0--33\%} & \multicolumn{2}{c}{33--67\%} & \multicolumn{2}{c|}{67--100\%} & 0--33\% & 33--67\% & 67--100\% & \multicolumn{2}{c}{0--33\%} & \multicolumn{2}{c}{33--67\%} & \multicolumn{2}{c|}{67--100\%} & 0--33\% & 33--67\% & 67--100\% \\
\midrule
HandOccNet~\citep{park2022handoccnet} & 5.67 & 5.47 & 6.63 & 6.43 & \phantom{0}7.78 & \phantom{0}\underline{7.53} & \na & \na & \na & 8.25 & 8.60 & 7.72 & 8.20 & 7.77 & 8.13 & \na & \na & \na \\
MobRecon~\citep{chen2022mobrecon} & 6.22 & 5.41 & 7.27 & 6.56 & \phantom{0}9.68 & \phantom{0}9.14 & \na & \na & \na & 8.22 & 8.62 & 8.09 & 8.55 & 7.86 & 8.26 & \na & \na & \na \\
H2ONet~\citep{xu2023h2onet} & 5.14 & 5.29 & 6.36 & 6.44 & \phantom{0}7.97 & \phantom{0}7.92 & \na & \na & \na & 8.12 & 8.59 & 8.03 & 8.64 & 7.84 & 8.48 & \na & \na & \na \\
SimpleHand~\citep{zhou2024simplehand} & 5.68 & 5.44 & 6.72 & 6.51 & 11.46 & 11.12 & \na & \na & \na & 8.62 & 9.02 & 8.32 & 8.76 & 8.87 & 9.12 & \na & \na & \na \\
\midrule
SemiHand~\citep{liu2021semihand} & 5.40 & 5.41 & 6.00 & 6.03 & \phantom{0}9.00 & \phantom{0}9.05 & 86.96 & 72.21 & 25.44 & 8.01 & 8.38 & 7.64 & 8.07 & 8.08 & 8.43 & 41.35 & \phantom{0}38.16 & 38.40 \\
HFL-Net~\citep{lin2023hfl} & 5.32 & 5.30 & 6.13 & 6.15 & \phantom{0}\underline{7.71} & \phantom{0}7.75 & 80.93 & 62.43 & 18.22 & 8.20 & 8.54 & 7.97 & 8.45 & 8.27 & 8.68 & 39.48 & \phantom{0}34.97 & 23.49 \\
UniHOPE~\citep{wang2025unihope} & 5.29 & 5.27 & 6.07 & 6.09 & \phantom{0}7.86 & \phantom{0}7.90 & 78.70 & 61.93 & 20.52 & 7.99 & 8.34 & 7.66 & 8.09 & 7.85 & 8.20 & 39.67 & \phantom{0}37.51 & 30.31 \\
\textbf{\method\ (ours)} & \underline{4.96} & \underline{4.92} & \textbf{5.18} & \textbf{5.17} & \phantom{0}\textbf{7.22} & \phantom{0}\textbf{7.15} & \underline{93.51} & \underline{94.41} & \textbf{72.65} & \underline{7.14} & \underline{7.59} & \underline{7.31} & \underline{7.84} & \underline{7.11} & \underline{7.52} & \underline{92.66} & \phantom{0}\underline{95.72} & \underline{90.55} \\
\textbf{\methodall\ (ours)} & \textbf{4.84} & \textbf{4.82} & \underline{5.30} & \underline{5.29} & \phantom{0}7.83 & \phantom{0}7.76 & \textbf{96.15} & \textbf{95.18} & \underline{68.82} & \textbf{4.82} & \textbf{4.85} & \textbf{4.69} & \textbf{4.75} & \textbf{5.26} & \textbf{5.29} & \textbf{99.96} & \textbf{100.00} & \textbf{99.90} \\
\bottomrule
\end{tabular}}
\end{table}

\vspace{1em}
\begin{wraptable}{r}{0.5\linewidth}
\vspace{-1.4em}
\centering
\caption{Object AUC under increasing object occlusion, including the object-pose estimators. \method\ degrades the least.}
\label{tab:objocc_full}
\par\vspace{-0.5em}
\resizebox{\linewidth}{!}{%
\begin{tabular}{l|ccc|ccc}
\toprule
& \multicolumn{3}{c|}{\textbf{Source: DexYCB}~\citep{chao2021dexycb}} & \multicolumn{3}{c}{\textbf{Target: HO3D}~\citep{hampali2020honnotate}} \\
\cmidrule(lr){2-4}\cmidrule(lr){5-7}
& \multicolumn{3}{c|}{Object Occlusion} & \multicolumn{3}{c}{Object Occlusion} \\
\cmidrule(lr){2-4}\cmidrule(lr){5-7}
\textbf{Method} & 0--33\% & 33--67\% & 67--100\% & 0--33\% & 33--67\% & 67--100\% \\
\midrule
MegaPose~\citep{labbe2022megapose} & 89.67 & 69.59 & 17.51 & \underline{96.94} & \underline{92.53} & \phantom{0}\underline{54.17} \\
GigaPose~\citep{nguyen2024gigapose} & 74.95 & 46.95 & 12.13 & 89.20 & 73.66 & \phantom{0}45.83 \\
FoundPose~\citep{ornek2024foundpose} & 73.22 & 40.89 & \phantom{0}7.49 & 92.61 & 79.95 & \phantom{0}12.50 \\
PoseGAM~\citep{chen2026posegam} & 76.37 & 32.41 & \phantom{0}9.61 & 96.85 & 20.58 & \phantom{0}\phantom{0}8.33 \\
\midrule
SemiHand~\citep{liu2021semihand} & 86.96 & 72.21 & 25.44 & 87.88 & 78.37 & \phantom{0}\phantom{0}8.33 \\
HFL-Net~\citep{lin2023hfl} & 80.93 & 62.43 & 18.22 & 83.03 & 64.22 & \phantom{0}16.67 \\
UniHOPE~\citep{wang2025unihope} & 78.70 & 61.93 & 20.52 & 81.33 & 63.56 & \phantom{0}20.83 \\
\textbf{\method\ (ours)} & \underline{93.51} & \underline{94.41} & \textbf{72.65} & 89.34 & 82.44 & \textbf{100.00} \\
\textbf{\methodall\ (ours)} & \textbf{96.15} & \textbf{95.18} & \underline{68.82} & \textbf{97.27} & \textbf{97.64} & \textbf{100.00} \\
\bottomrule
\end{tabular}}
\vspace{-1em}
\end{wraptable}
\noindent This also holds against the object-pose estimators in \mbox{Table~\ref{tab:objocc_full}}, which match rendered views of the mesh to the image.
On DexYCB their AUC falls from between 73.22 and 89.67 in the first bin to between 7.49 and 17.51 in the last, below the joint estimators.
The grasp hides the object regions that these views rely on.
Within \mbox{\method}, the same PoseGAM~\citep{chen2026posegam} feature degrades far less with 72.65 against 9.61 in the last bin.
MegaPose~\citep{labbe2022megapose} exceeds \method\ only in the two less occluded HO3D bins with 96.94 and 92.53 against 89.34 and 82.44.
Joint reasoning about the object and the grasping hand thus compensates for the evidence hidden by the grasp.
\subsection{Ablation of Foundation Features}
\label{app:ablation}
We measure the contribution of the two frozen foundation features~\citep{chen2026posegam, pavlakos2024hamer} by removing each from the DexYCB-s0 model and retraining the first stage with all else unchanged.
Each variant is evaluated in domain and on the transfer targets without retraining.
As shown in Table~\ref{tab:abl_feat}, removing the object-pose feature of the given mesh lowers the average object AUC from 85.65 to 77.00.
The ablated model still places the object at the grasping hand but largely loses its rotation, as its median rotation error rises from at most $27^\circ$ to at least $50^\circ$ on every dataset.
On OakInk the closest-point ADD-S hides this loss with 94.08 against 93.29, but the plain ADD of Table~\ref{tab:strictobj} reveals it with a drop from 56.40 to 45.97.
The MJE changes by at most 1.23, so this feature serves the object alone.
Removing the hand-pose backbone instead raises the average MJE from 18.00 to 30.71.
The increase is larger on every target than in domain, reaching 21.87 on HO3D against 7.21 on DexYCB.
The object AUC also drops by 2.73 to 4.47 points, as the object is placed relative to the predicted hand.
Since \method\ uses only the backbone features of the hand-pose model and learns the hand pose on the joint task, these features mainly provide a transferable prior on hand appearance.
The object-pose feature thus recovers the object rotation, whereas the hand-pose backbone drives the hand accuracy and its transfer across datasets.
\begin{table}[!h]
\centering
\caption{Ablation of the frozen foundation features of the DexYCB-s0 model in domain and on the transfer targets without retraining. \mbox{\emph{w/o Object Feature}} removes the object-pose feature of Eqn.~\eqref{eq:inject}, and \mbox{\emph{w/o Hand Feature}} removes the hand-pose backbone from $E_h$ of Eqn.~\eqref{eq:fuse}. Average is the mean over the four datasets. The object feature mainly drives the object AUC, and the hand feature the MJE.}
\label{tab:abl_feat}
\par\vspace{-0.5em}
\resizebox{\linewidth}{!}{%
\begin{tabular}{l|cc|cc|cc|cc|cc}
\toprule
& \multicolumn{2}{c|}{\textbf{DexYCB}~\citep{chao2021dexycb}} & \multicolumn{2}{c|}{\textbf{HO3D}~\citep{hampali2020honnotate}} & \multicolumn{2}{c|}{\textbf{OakInk}~\citep{yang2022oakink}} & \multicolumn{2}{c|}{\textbf{HOI4D}~\citep{liu2022hoi4d}} & \multicolumn{2}{c}{\textbf{Average}} \\
\cmidrule(lr){2-3}\cmidrule(lr){4-5}\cmidrule(lr){6-7}\cmidrule(lr){8-9}\cmidrule(lr){10-11}
\textbf{Method} & AUC ($\uparrow$) & MJE ($\downarrow$) & AUC ($\uparrow$) & MJE ($\downarrow$) & AUC ($\uparrow$) & MJE ($\downarrow$) & AUC ($\uparrow$) & MJE ($\downarrow$) & AUC ($\uparrow$) & MJE ($\downarrow$) \\
\midrule
Ours w/o Object Feature~\citep{chen2026posegam} & 76.45 & \underline{10.77} & 78.41 & \underline{23.45} & \textbf{94.08} & \underline{16.16} & 59.06 & \underline{23.90} & 77.00 & \underline{18.57} \\
Ours w/o Hand Feature~\citep{pavlakos2024hamer} & \underline{89.61} & 17.33 & \underline{86.02} & 44.09 & 89.09 & 27.13 & \underline{63.21} & 34.29 & \underline{81.98} & 30.71 \\
\method\ (ours) & \textbf{92.88} & \textbf{10.12} & \textbf{88.75} & \textbf{22.22} & \underline{93.29} & \textbf{16.02} & \textbf{67.68} & \textbf{23.64} & \textbf{85.65} & \textbf{18.00} \\
\bottomrule
\end{tabular}}
\end{table}

\FloatBarrier
\section{Limitations and Future Work}
\label{app:limitations}
\textbf{Limitations.}\quad
While \method\ generalizes well to unseen objects, several limitations warrant further investigation.
First, \method\ encodes the object from its geometry and thus requires an object mesh as input.
Second, its accuracy may drop on objects and scenes far from the training data (\eg, articulated, deformable, or unusually shaped objects).
Third, \method\ inherits the fixed input resolution and pretraining biases of its frozen foundation encoders.
Addressing these limitations would further broaden the applicability of \method\ to hand-object interactions in the wild.

\vspace{0.5em}\noindent\textbf{Future work.}\quad
\method\ opens up several promising avenues for future research.
First, estimating the object shape jointly with the poses would remove the need for an input mesh.
Second, initializing a physics simulator with the reconstructed hands, object, and fine contact could yield stable and penetration-free grasps that in turn provide physical supervision for training.
Third, extending \method\ from single images to videos would resolve single-frame depth and occlusion ambiguities and reveal how contact forms and releases over time.
Progress in these directions would bring \method\ closer to mesh-free, physically plausible, and temporally consistent reconstruction of everyday hand-object interactions.

\end{document}